\documentclass{article}

\usepackage{microtype}
\usepackage[accepted,nohyperref]{mlsys2021}

\mlsystitlerunning{ByzShield: An Efficient and Robust System for Distributed Training}
\usepackage{calc}

\usepackage{hyperref} % I did not use MLSys hyperref, it was giving errors
\usepackage{graphics,color,wrapfig}

\usepackage{url}

\usepackage{breakurl}

\usepackage{color}
\usepackage{amsthm}
\usepackage{amsmath,amssymb}
\usepackage{amsfonts}
\usepackage{array}
\newcolumntype{P}[1]{>{\centering\arraybackslash}m{#1}}
\usepackage{multirow}
\usepackage{graphicx}
\usepackage{standalone}
\usepackage{booktabs}
\usepackage{algpseudocode} % For BYZSHIELD this line was uncommented
\usepackage{algorithm2e} % For BYZSHIELD this line was uncommented
\let\oldnl\nl% Store \nl in \oldnl
\newcommand{\nonl}{\renewcommand{\nl}{\let\nl\oldnl}}% Remove line number for one line
\usepackage{makecell}
\usepackage[bottom]{footmisc}
\usepackage{enumitem}
\theoremstyle{definition}
\usepackage{tikz}
\usepackage{xcolor}

\usepackage[justification=centering]{caption} % Figures caption
\usepackage{subcaption}

\usepackage{mathtools}

\newcommand\bigzero{\makebox(0,0){\text{\huge0}}}

\usepackage{bbm}

\makeatletter
\renewcommand{\@algocf@capt@plain}{above}% formerly {bottom}
\makeatother
\RestyleAlgo{ruled}
\LinesNumbered

\renewcommand{\em}{\it}

\newcommand{\calN}{\mathcal{N}}

\newcommand{\calU}{\mathcal{U}}
\newcommand{\calF}{\mathcal{F}}
\newcommand{\calE}{\mathcal{E}}

\newcommand{\bfG}{\mathbf{G}}

\newtheorem{lemma}{Lemma}

\newtheorem{definition}{Definition}
\newtheorem{claim}{Claim}
\newtheorem{example}{Example}

\newtheorem{note}{Note}

{\begin{list}{}%
		{\setlength{\leftmargin}{#1}}%
		\item[]%
	}
	{\end{list}}

\newcounter{relctr} %% <- counter for relations
\everydisplay\expandafter{\the\everydisplay\setcounter{relctr}{0}} %% <- reset every eq
\AtBeginDocument{} %% <- store original definition
\DeclareCaptionSubType * [alph]{table}
\newcommand\defeq{\stackrel{\mathclap{\normalfont\mbox{def}}}{=}}

\begin{document}

\twocolumn[
\mlsystitle{ByzShield: An Efficient and Robust System for Distributed Training}

% unless you've provided the [accepted] option to the mlsys2021


\mlsyssetsymbol{equal}{*}

\begin{mlsysauthorlist}
\mlsysauthor{Konstantinos Konstantinidis}{isu_ecpe}
\mlsysauthor{Aditya Ramamoorthy}{isu_ecpe}
\end{mlsysauthorlist}

\mlsysaffiliation{isu_ecpe}{Department of Electrical and Computer Engineering, Iowa State University, Ames, IA 50011 USA}

\mlsyscorrespondingauthor{Konstantinos Konstantinidis}{kostas@iastate.edu}
\mlsyscorrespondingauthor{Aditya Ramamoorthy}{adityar@iastate.edu}

\mlsyskeywords{TBA}

\vskip 0.3in

\begin{abstract}
%The training of large scale models on distributed clusters is vulnerable to attack when some workers behave in a Byzantine fashion. Such behavior is commonly attributed to malicious attacks (and/or system failures) in which a subset of the workers can exhibit abnormal behavior and return arbitrary results to the \emph{parameter server} (PS). A plethora of existing papers propose \emph{robust aggregation} and/or computation \emph{redundancy} to alleviate the effect of distorted gradients. We show that these schemes can be adversely impacted in \emph{omniscient} scenarios where an adversary knows the task assignment and can appropriately attack workers to induce maximal damage. We propose a set of algorithms, named \emph{ByzShield},  to carefully assign gradients to workers such that the worst-case distortion is minimized. Our assignment is based on matrices known as \emph{Latin squares} as well as \emph{Ramanujan} bipartite graphs with desired expansion properties and has a similar computation load per worker as prior work. We analytically prove the Byzantine resilience guarantees we can achieve in the worst case and extensively evaluate the system on various large-scale deep learning tasks. We also provide exact worst-case results for some regimes. The main metric for our experiments is the accuracy of the training for which we demonstrate significant improvements compared to many state-of-the-art approaches. A secondary metric we analyze is the convergence time.

Training of large scale models on distributed clusters is a critical component of the machine learning pipeline. However, this training can easily be made to fail if some workers behave in an adversarial (\emph{Byzantine}) fashion whereby they return arbitrary results to the \emph{parameter server} (PS). A plethora of existing papers consider a variety of attack models and propose \emph{robust aggregation} and/or computational \emph{redundancy} to alleviate the effects of these attacks. In this work we consider an \emph{omniscient} attack model where the adversary has full knowledge about the gradient computation assignments of the workers and can choose to attack (up to) any $q$ out of $K$ worker nodes to induce maximal damage. Our redundancy-based method \emph{ByzShield} leverages the properties of bipartite expander graphs for the assignment of tasks to workers; this helps to effectively mitigate the effect of the Byzantine behavior. Specifically, we demonstrate an upper bound on the worst case fraction of corrupted gradients based on the eigenvalues of our constructions which are based on \emph{mutually orthogonal Latin squares} and \emph{Ramanujan} graphs. Our numerical experiments indicate over a 36\% reduction on average in the fraction of corrupted gradients compared to the state of the art. Likewise, our experiments on training followed by image classification on the CIFAR-10 dataset show that ByzShield has on average a 20\% advantage in accuracy under the most sophisticated attacks. ByzShield also tolerates a much larger fraction of adversarial nodes compared to prior work.
\end{abstract}
]



\printAffiliationsAndNotice{}  % leave blank if no need to mention equal contribution
%\printAffiliationsAndNotice{\mlsysEqualContribution} % otherwise use the standard text.

\section{Introduction}
%\aditya{will need work, allocate 2 pages}
The explosive growth of machine learning applications have necessitated the routine training of large scale ML models in a variety of domains. Owing to computational and space constraints this training is typically run on distributed clusters. A common architecture consists of a single central server (\emph{parameter server} or PS) which maintains a global copy of the model and several worker machines. 
% The structure of the network is agreed between all machines beforehand.
The workers compute gradients of the loss function with respect to the parameters being optimized. The results are returned to the PS which \emph{aggregates} them and updates the model. A new copy of the model is broadcasted from the PS to the workers and the above procedure is repeated until convergence. TensorFlow \cite{tensorflow}, MXNet \cite{mxnet}, CNTK \cite{CNTK} and PyTorch \cite{pytorch} are all examples of this architecture.

%%%%%%%%%%%%%%%%%%%%%%%%%%%%%%%%%%%%%%%%%%%%%%%%%%%%%%%%%%%%%%%%%%%%%%%%%%%%%%%%%%%%%%%%%%%%%%%%%%%%%%%%%%%%%%%%%%%%%%%%%%%%%%%%%%%%%%
% TO BE INCLUDED
%%%%%%%%%%%%%%%%%%%%%%%%%%%%%%%%%%%%%%%%%%%%%%%%%%%%%%%%%%%%%%%%%%%%%%%%%%%%%%%%%%%%%%%%%%%%%%%%%%%%%%%%%%%%%%%%%%%%%%%%%%%%%%%%%%%%%%
% Privacy is another motivation for distributed ML schemes, e.g., when users want to keep their data locally and collaborate via a PS to contribute to some global machine learning task \cite{abadi_diff_privacy, shokri_privacy}. This is a desired property in \emph{federated learning} in which work is distributed among a large number of compute nodes without centralized training data \cite{jakub_fed_learning, jakub_fed_optimization, bonawitz_fed_learning}.
%%%%%%%%%%%%%%%%%%%%%%%%%%%%%%%%%%%%%%%%%%%%%%%%%%%%%%%%%%%%%%%%%%%%%%%%%%%%%%%%%%%%%%%%%%%%%%%%%%%%%%%%%%%%%%%%%%%%%%%%%%%%%%%%%%%%%%

A major challenge with such setups as they scale is the fact that the system is vulnerable to adversarial attacks by the computing nodes (i.e., \emph{Byzantine} behavior), or by failures/crashes thereof.\footnote{In this work, we assume that the PS is reliable.}  As a result, the corresponding gradients returned to the PS are potentially unreliable for use in training and model updates. Achieving robustness in the face of Byzantine node behavior (which can cause gradient-based methods to converge slowly or to diverge) is of crucial importance and has attracted significant attention \cite{gunawi_fail_slow, large_scale_deep_ng, kotla_zyzzyva}. 

One class of methods investigates techniques for suitably aggregating the results from the workers. \cite{blanchard_krum} establishes that no \emph{linear aggregation} method (such as averaging) can be robust even to a single faulty worker. \emph{Robust aggregation} has been proposed as an alternative. Majority voting, geometric median and squared-distance-based techniques fall into this category \cite{aggregathor, ramchandran_saddle_point, ramchandran_optimal_rates, cong_generalized_sgd, blanchard_krum, yudong_lilisu}. The benefit of such approaches is their robustness guarantees. However, they have serious weaknesses. First, they are very limited in terms of the fraction of Byzantine nodes they can withstand. Moreover, their complexity often scales quadratically  with the number of workers \cite{cong_generalized_sgd}. Finally, their theoretical guarantees are insufficient to establish convergence and require strict assumptions such as convexity of the loss function. %\aditya{convexity of what is assumed?}

Another category of defenses is based on \emph{redundancy} and seeks resilience to Byzantines by replicating the gradient computations such that each of them is processed by more than one machine \cite{lagrange_cdc, detox, draco, data_encoding}. Even though this approach requires more computation load, it comes with stronger guarantees of correcting the erroneous gradients. Existing techniques are sometimes combined with robust aggregation \cite{detox}. The main drawback of recent work in this category is that the training can be easily disrupted by a powerful omniscient adversary who has full control of a subset of the nodes and can mount judicious attacks.

%\subsection{Contributions}
%In this work, we propose a set of algorithms that can be classified as redundancy-based. Our method, named \emph{ByzShield}, splits the batch of an update into multiple sets and assigns copies of them to workers such that the number of corrupted gradients is minimized. We work with \emph{omniscient} setups where the attacker knows the placement of the replicated files and chooses which devices to attack adversarially.
%in an adversarial manner and test our scheme under such worst-case attacks.

%Our placement uses \emph{Latin squares} and \emph{Ramanujan} bigraphs. In both cases, we interpret the assignment as a problem of matching nodes in a bipartite graph and establish connections between the expansion properties of the graph and the achieved robustness.

%Our theoretical analysis proves that our method achieves a much lower fraction of corrupted gradients compared to other methods \cite{detox}; this is supported by numerical simulations. Our experimental analysis involves large scale training for image classification problems. The results show that ByzShield has on average a 20\% advantage in terms of accuracy under the most sophisticated attacks.

%We also propose simple attacks (for the omniscient case) which make existing algorithms \cite{detox} fail and produce non-convergent models. We show that ByzShield can converge to accurate models even when the number of Byzantine workers is large; in these cases, state-of-the-art methods either diverge or are not applicable \cite{aggregathor, detox, bulyan}.

%AR: Putting in new section
\subsection{Contributions}
In this work, we propose our redundancy-based method \emph{ByzShield} that provides robustness in the face of a significantly stronger attack model than has been considered in prior work. In particular, we consider an omniscient adversary who has full knowledge of the tasks assigned to the workers. The adversary is allowed to pick any subset of $q$ worker nodes to attack at each iteration of the algorithm. Our model allows for collusion among the Byzantine workers to inflict maximum damage on the training process.

Our work demonstrates an interesting link between achieving robustness and spectral graph theory. We show that using optimal expander graphs (constructed from \emph{mutually orthogonal Latin squares} and \emph{Ramanujan} graphs) allows us to assign redundant tasks to workers in such a way that the adversary's ability to corrupt the gradients is significantly reduced as compared to the prior state of the art. 

Our theoretical analysis shows that our method achieves a much lower fraction of corrupted gradients compared to other methods \cite{detox}; this is supported by numerical simulations which indicate over a 36\% reduction on average in the fraction of corrupted gradients. Our experimental results on large scale training for image classification (CIFAR-10) show that ByzShield has on average a 20\% advantage in accuracy under the most sophisticated attacks.

Finally, we demonstrate the fragility of existing schemes \cite{detox} to simple attacks under our attack model when the number of Byzantine workers is large. In these cases, prior methods either diverge or are not applicable \cite{aggregathor, detox, bulyan}, whereas ByzShield converges to high accuracy.

\subsection{Related Work}
Prior work in this area considers adversaries with a wide range of powers. For instance, DETOX and DRACO consider adversaries that are only able to attack a random set of workers. In our case, an adversary possesses full knowledge of the task assignment and attacked workers can collude.
%\aditya{one or two lines}
Prior work which considers the same adversarial model as ours includes \cite{alie, cong_generalized_sgd, ramchandran_optimal_rates, bulyan, auror} . %\aditya{check for accuracy}

%can control the devices themselves. This is a strong attack formulation since it allows for a careful targeting of specific workers to induce maximal damage. 
%Similar attacks are considered in \cite{cong_generalized_sgd, ramchandran_optimal_rates, bulyan, auror}.

One of the most popular robust aggregation techniques is known as \emph{mean-around-median} or \emph{trimmed mean} \cite{cong_generalized_sgd, ramchandran_optimal_rates, bulyan}. It handles each dimension of the gradient separately and returns the average of a subset of the values that are closest to the median.
%; these values are chosen based on some criterion which depends on the implementation. 
\emph{Auror} introduced in \cite{auror} is a variant of trimmed median which partitions the values of each dimension into two clusters using \emph{k-means} and discards the smaller cluster if the distance between the two exceeds a threshold; the values of the larger cluster are then averaged.
%(median must be among them). 
\emph{signSGD} in \cite{SIGNSGD} transmits only the sign of the gradient vectors from the workers to the PS and exploits majority voting to decide the overall update; this practice improves communication time and denies any individual worker too much effect on the update.

\emph{Krum} in \cite{blanchard_krum} 
%takes a different approach and tries to choose 
chooses a single honest worker for the next model update.
%, discarding the data from the rest of them. 
The chosen gradient is the one closest to its $k\in\mathbb{N}$ nearest neighbors.
% measured by Euclidean distance. 
The authors recognized in later work \cite{bulyan} that Krum may converge to an \emph{ineffectual} model in the landscape of non-convex high dimensional problems.
%, such as in neural networks. 
They showed that a large adversarial change to a single parameter with a minor impact on the $L^p$ norm can make the model ineffective. They present an alternative defense called \emph{Bulyan} to oppose such attacks. 
%The algorithm works in two stages. In the first part, a \emph{selection set} of potentially benign values is iteratively constructed. In the second part, a variant of trimmed mean is applied to the selection set. 
Nevertheless, if $K$ machines are used, Bulyan is designed to defend only up to $(K-3)/4$ fraction of corrupted workers. 
Similarly, the method of \cite{yudong_lilisu} is based on the \emph{geometric median of means}. 

%The benefit of such approaches is the robustness guarantees they provide. 
%However, the cost of computing the geometric median can be prohibitively high compared to gradient computations.
%However, the number of adversarial nodes they can tolerate is fairly limited. Furthermore, convergence proofs in these cases require strict assumptions such as convexity which limits their flexibility.

%The other major category of defenses is based on redundancy of the gradient tasks where the batch is split into smaller sets of samples and each set is assigned to multiple machines. 
In the area of redundancy-based methods, \emph{DRACO} in \cite{draco} uses a simple \emph{Fractional Repetition Code} (FRC) (that operates by grouping workers) and the cyclic repetition code introduced in \cite{dimakis_cyclic_mds, tandon_gradient} to ensure robustness; majority vote and Fourier decoders try to alleviate the adversarial effects. Their work ensures exact recovery (as if the system had no adversaries) with $q$ Byzantine nodes, when each task is replicated $r \geq 2q+1$ times; the bound is information-theoretic minimum and DRACO is not applicable if it is violated. Nonetheless, this requirement is very restrictive for the typical assumption that up to half of the workers can be Byzantine. %In this adverse scenario, each worker in DRACO will have to process the entire data set which contradicts the objective of distributed computing; communication bottlenecks and storage needs will also increase accordingly.

%\emph{DETOX} in \cite{detox} extends the work of \cite{draco} and uses the FRC to assign the gradients. It performs multiple stages of aggregation to gradually filter the adversarial values. The first stage involves majority voting while the following stages perform robust aggregation, as discussed before.

%AR: Rewording here
\emph{DETOX} in \cite{detox} extends the work of \cite{draco} and uses a simple grouping strategy to assign the gradients. It performs multiple stages of aggregation to gradually filter the adversarial values. The first stage involves majority voting while the following stages perform robust aggregation, as discussed before.
% (\emph{cf.} geometric median \cite{yudong_lilisu, ramchandran_optimal_rates}, Bulyan \cite{bulyan}, \emph{Multi-Krum} \cite{aggregathor}). 
Unlike DRACO, the authors do not seek exact recovery hence the minimum requirement in $r$ is small. However, the theoretical resilience guarantees that DETOX provides depend heavily on a ``random assignment'' of tasks to workers and on ``random choice'' of the adversarial workers. Furthermore, their theoretical results hold when the fraction of Byzantines is less than $1/40$. Under these assumptions, they show that on average a small percentage of the gradient computations will be distorted.

%As mentioned earlier, we consider attacks where one can decide which machines will fail. To that end, we give simple attacks that exploit the placement policies of \cite{draco,detox} and effectively distort the maximal possible number of majority votes given a value of $q$. We then show that for various malignant gradient manipulation methods our attack can make the training diverge. Instead, ByzShield provides a more systematic choice of the workers to execute a given task that is significantly more robust to such cases.

\section{Parallel Training Formulation}
\label{sec:formulation}
The formulation we discuss is standard in distributed deep learning. Assume that the loss function of the $i$-th sample is $l_i(\mathbf{w})$, where $\mathbf{w}\in\mathbb{R}^d$ is the parameter set of the model. We seek to minimize the \emph{empirical risk} of the dataset, i.e., 
\begin{equation*}
	\min_{\mathbf{w}}L(\mathbf{w})=\min_{\mathbf{w}}\frac{1}{n}\sum\limits_{i=1}^nl_i(\mathbf{w})
\end{equation*}
where $n$ is the size of the dataset. This is typically solved iteratively via \emph{mini-batch Stochastic Gradient Descent} (SGD) over distributed clusters. Initially, the parameters $\mathbf{w}$ are randomly set to $\mathbf{w}_0$ (in general, we will denote the state of the model at the end of iteration $t$ with $\mathbf{w}_t$). Following this, a randomly chosen \emph{batch} $B_t$ of $b$ samples is used to perform the update in the $t$-th iteration. Thus, %If we use $\eta_t$ as the learning rate for the $t$-th iteration, then the vanilla update would be
\begin{equation}
	\label{eq:vanilla_sgd_update}
	\mathbf{w}_{t+1}=\mathbf{w}_{t}-\eta_t\frac{1}{|B_t|}\sum\limits_{i\in B_t}\nabla l_i(\mathbf{w}_t)
\end{equation}
where $\eta_t$ denotes the learning rate of the $t$-th iteration. In this work, we focus on \emph{synchronous} SGD in which the PS waits for all workers to return before performing an update.

%\DecMargin{1em}
%\begin{algorithm}[!t]
%	%\LinesNotNumbered
%	%\SetAlgoNoLine
%	%\KwIn{Workers $\{U_0,U_1\dots,U_{K-1}\}$, parameter server (PS), batch of current update $B_t$, batch size $b$, computation load $l$, redundancy factor $r\leq l-1$.}
%	\KwIn{Worker set $\calU = \{U_0,U_1\dots,U_{K-1}\}$, \newline bipartite graph $\bfG = (\calU \cup \calF_t, \calE)$.}
%	
%	{
%		\abovedisplayskip=0pt
%		\belowdisplayskip=0pt
%		%The PS constructs a  $B_t$ into $l^2$ disjoint files of $b/l^2$ samples each $$B_t=\left\{B_{t,i}: i=0,1,\dots,f-1\right\}.$$\\
%		The PS divides $B_t$ into $f$ disjoint equal-sized files denoted $$\calF_t=\left\{B_{t,i}: i=0,1,\dots,f-1\right\}.$$\\
%		%Let $\calU = \{U_0,U_1\dots,U_{K-1}\}$ and $\calF = \{B_{t,0},B_{t,1}\dots,B_{t,f-1}\}$.\\
%		%The PS constructs a $(l,r)$ biregular bipartite graph from $\calU$ to $\calF_t$.\\
%		\For{each edge $(x,y) \in \calE$ s.t. $x \in \calU$, $y \in \calF_t$}{
%			PS assigns file $y$ to worker $x$.
%		}
%	}
%	\caption{Proposed file assignment based on bipartite graphs.}
%\end{algorithm}

Within the cluster of $K+1$ servers the PS updates the global model after receiving computations from the workers. It also stores the dataset and the model and coordinates the protocol. The remaining servers, denoted $U_0,\dots,U_{K-1}$, are the workers computing gradients on subsets of the batch. 
%\aditya{remove Algo 1. Only keep one Algo that includes bipartite graph that specifies the placement}

\textbf{Worker Assignment}: A given batch $B_t$ is split into $f$ disjoint sets (or \emph{files}) denoted $B_{t,i}$, for $i = 0, \dots, f-1$. These files are then assigned to the workers according to a bipartite graph $\bfG = (\calU \cup \calF_t, \calE)$, where $\calU =  \{U_0,U_1\dots,U_{K-1}\}$ and $\calF_t = \left\{B_{t,i}: i=0,1,\dots,f-1\right\}$ denote the workers and the files, respectively. An edge exists between $u \in \calU$  and $v \in \calF_t$ if worker $u$ is assigned file $v$. Any given file is assigned to $r>1$ workers ($r$ is called the \emph{replication factor}); this allows for protection against Byzantine behavior. It follows that each worker is responsible for $l = fr/K$ files; $l$ is called the \emph{computational load}. We let $\calN(S)$ denote the set of neighbors of a subset $S$ of the nodes. Thus, $\calN(U_j)$ is the set of files assigned to worker $U_j$ and $\calN(B_{t,i})$ is the set of workers that are assigned file $B_{t,i}$.%In this work, we propose establish a connection between certain properties of such bipartite graphs and explicitly construct them.

%\aditya{Your notation convention for the sets is not standard and is likely to cause confusion. I would suggest you avoid it}

{\bf Byzantine Attack Model}: We consider a model where $q$ workers operate in a Byzantine fashion, i.e., they can return arbitrary values to the PS. Our attack setup is \emph{omniscient} where every worker knows the data assignment of all other participants and the model at {\it every} iteration; the choice of the Byzantine workers can also change across iterations. Furthermore, the Byzantine machines can collude to induce maximal damage. We will suppose that the fraction of Byzantine workers $\epsilon=q/K$ is less than $1/2$. We emphasize that this is a stronger adversary model than those considered in other redundancy-based work \cite{detox, draco}.
%\aditya{add citations}.
%For example, \emph{DETOX} in \cite{detox} considers $K/r$ groups of workers and assigns the exact same set of files to all participants in a group. The set of Byzantine workers is chosen randomly at each iteration. However, as we will see, if the Byzantine workers are chosen adversarially, then the performance of their scheme can be much worse. %This can crucially hurt their scheme if a stronger attack model allows the adversary to choose which workers to affect; e.g., an attacker can choose exactly $r'$ Byzantines from as many groups as possible such that majority-based methods will fail.
Let $\hat{\mathbf{g}}_{t,i}^{(j)}$ be the value returned from the $j$-th worker to the PS for assigned file $B_{t,i}$. Then,
\begin{equation}
	\label{eq:gradient_hat}
	\hat{\mathbf{g}}_{t,i}^{(j)} = \left\{
	\begin{array}{ll}
		\mathbf{g}_{t,i} & \text{ if } U_j \text{ is honest},\\
		\mathbf{x} & \text{otherwise} \\
	\end{array} 
	\right.
\end{equation}
where $\mathbf{g}_{t,i}$ is the sum of the loss gradients on all samples in file $B_{t,i}$, i.e.,
\begin{equation*}
\mathbf{g}_{t,i} = \sum\limits_{j\in B_{t,i}}\nabla l_j(\mathbf{w}_t) 
\end{equation*}
and $\mathbf{x}$ is any arbitrary vector in $\mathbb{R}^d$.

\begin{figure}[t]
	\centering
	\includegraphics[scale=1.1]{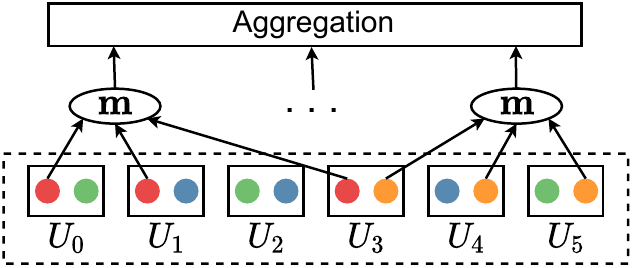}
	\caption{Aggregation of gradients on a training cluster.}
	\label{fig:aggregation_general}
	\vspace{-0.1in}
\end{figure}

{\bf Training Protocol}: %The proposed protocols assign replicas of each sample batch to multiple workers. This practice introduces computation redundancy but yields significant advantages in terms of resilience to Byzantines attacks or failures. Let us consider a batch $B_t$ of a single iteration. Instead of assigning each file $B_{t,i}$ to one worker, we will assign it to a group of $r>1$ workers, whose indices are denoted with $W_{t,i}$ (let us use the term \emph{replication} for $r$). 
The training protocol followed in our work is similar to the one introduced in \cite{detox}. Following the assignment of files, each worker $U_j$ computes the gradients of the samples in $\calN(U_j)$ and sends them to the PS. Subsequently, the PS will perform a majority vote across the computations of each file. Recall that each file has been processed by $r$ workers. %Let the workers that have processed $B_{t,i}$ be $U_{j_1},U_{j_2},\dots,U_{j_r}$. 
For each such file $B_{t,i}$, the PS decides a majority value $\mathbf{m}_i$

\begin{equation}
	\label{eq:basic_majority}
%	\begin{array}{ll}
		\mathbf{m}_i \defeq \mathrm{majority}\left\{\hat{\mathbf{g}}_{t,i}^{(j)}: U_j \in \calN(B_{t,i})\right\}.
		% &=\mathrm{majority}\left\{\hat{\mathbf{g}}_{t,i}^{(j)}: j\in\{W_{t,i}\}\right\}.
%	\end{array}
\end{equation}

%\begin{equation}
%	\label{eq:basic_majority}
%	\begin{array}{ll}
%		\mathbf{m}_i &\coloneqq %\mathrm{majority}\left\{\hat{\mathbf{g}}_{t,i}^{(j)}: %j\in\{j_1,j_2,\dots,j_r\}\right\}
		% &=\mathrm{majority}\left\{\hat{\mathbf{g}}_{t,i}^{(j)}: j\in\{W_{t,i}\}\right\}.
%	\end{array}
%\end{equation}

%To distort at least one gradient we need $q\geq \lceil \frac{r}{2} \rceil = \frac{r+1}{2}$ for odd $r$. 
In our implementation, the majority function picks out the gradient that appears the maximum number of times. We ensure that there are no floating point precision issues with this method, i.e., all honest workers assigned to $B_{t,i}$ will return the exact same gradient. However, potential precision issues can easily be handled by deciding the majority by defining appropriate clusters amongst the returned gradients.

%In our implementation, we guarantee that there are no floating precision issues as long as all workers use the same compiler. Hence, all honest workers assigned to $B_{t,i}$ will return the exact same gradient and the majority aggregation looks for exact matches in order to count the occurrences of each vector. Alternatively, an approximate rule can be used in setups where there may be precision errors.

Assume that $r$ is odd (needed in order to avoid breaking ties) and let $r'=\frac{r+1}{2}$. %For tie-braking purposes we will consider only odd values of $r$.
Under the majority rule in Eq. \eqref{eq:basic_majority}, the gradient on a file is distorted only if at least $r'$ of the computations are performed by Byzantine workers. Following the majority vote, the gradients go through an aggregation step. One demonstration of this procedure is in Figure \ref{fig:aggregation_general}. There are $K=6$ machines and $f=4$ distinct files (represented by colored circles) replicated $r=3$ times. Each of the workers is assigned to two files and computes the sum of gradients (or an adversarial vector) on each file. At the next step, all returned values for the red file will be evaluated by a majority vote function (``$\mathbf{m}$" ellipses) on the PS which decides a single output value; a similar voting is done for the other 3 files. After the majority voting process, the ``winning" gradients $\mathbf{m}_i$, $i=0,1,\dots,f-1$ can be further filtered by some aggregation function; in ByzShield we chose to apply coordinate-wise median. The entire procedure is described in Algorithm \ref{alg:main_algorithm}.

%\aditya{provide some details here. Be clear on what is done in each step. Also need an appriate figure.}

{\bf Figures of Merit}: Firstly, we consider (i) the fraction of the estimated gradients $\mathbf{m}_i$ which are potentially erroneous. % and/or the norm of the error between the true and the computed gradient. 
However, the end goal is to understand how the presence of Byzantine workers affects the overall training of the model under the different schemes. Towards this end, we will also measure the (ii) accuracy of the trained model on classification tasks. Finally, robustness to adversaries comes at the cost of increased training time. Thus, our third metric is (iii) overall time taken to fit the model.
Asymptotic complexity (iv) is briefly addressed in the Appendix Section \ref{supplementary:asymptotic}.

%One typical measure of robustness is the fraction of the estimated gradients $\mathbf{m}_i$ which are potentially erroneous and/or the norm of the error between the true and the computed gradient. However, the end goal is to understand how the presence of Byzantine workers affects the overall training of the model under the different schemes. Towards this end, we will also measure the (i) accuracy of the trained model on classifying the test dataset, and (ii) the overall time taken to fit the model.

%Our end goal is to show that the proposed placement policy can effectively reduce the impact of Byzantine servers on the model's training accuracy and convergence.  

%\DecMargin{1em}
\begin{algorithm}[!t]
	%\LinesNotNumbered
	%\SetAlgoNoLine
	\KwIn{
		Data set of $n$ samples, batch size $b$, computation load $l$, redundancy $r$, \newline number of files $f$, maximum iterations $T$, assignment graph $\bfG$. %learning rates $\left\{\eta_t: t=0,1,\dots\right\}$.
	}
	{
		\abovedisplayskip=0pt
		\belowdisplayskip=0pt
		The PS randomly initializes model's parameters to $\mathbf{w}_0$.\\
		\For{$t = 1$ to $T$}{
			PS chooses a random batch $B_t\subseteq\{1,2,\dots,n\}$ of $b$ samples, partitions it into files and assigns them to workers based on graph $\bfG$. It then transmits model $\mathbf{w}_t$ to all workers.\\
			\For{each worker $U_j$}{
%				Let $X_{t,j} = \{i_1,i_2,\dots,i_l\}$ be the files of $U_j$.\\
				\eIf{$U_j$ is honest}{
%					$U_j$ computes the gradients of all files indexed in $X_{t,j}$.\\
					\For{each file $i \in \calN(U_j)$}{
						$U_j$ computes the sum of gradients $$\hat{\mathbf{g}}_{t,i}^{(j)}=\sum\limits_{k\in B_{t,i}}\nabla l_k(\mathbf{w}_t).$$
					}
				}{
					$U_j$ constructs adversarial vectors $$\hat{\mathbf{g}}_{t,i_1}^{(j)},\hat{\mathbf{g}}_{t,i_2}^{(j)},\dots,\hat{\mathbf{g}}_{t,i_l}^{(j)}.$$ %depending on the failure/attack model.
				}
				$U_j$ returns $\hat{\mathbf{g}}_{t,i_1}^{(j)},\hat{\mathbf{g}}_{t,i_2}^{(j)},\dots,\hat{\mathbf{g}}_{t,i_l}^{(j)}$ to the PS.
			}
			\For{$i=0$ to $f-1$}{
				\label{alg_step:main_maj_vote_start}
				PS determines the $r$ workers in $\calN(B_{t,i})$ which have processed $B_{t,i}$ and computes $$\mathbf{m}_i = \mathrm{majority}\left\{\hat{\mathbf{g}}_{t,i}^{(j)}: j \in \calN(B_{t,i})\}\right\}.$$
			}
			\label{alg_step:main_maj_vote_end}
			PS updates the model via
			\begin{equation*}
				\mathbf{w}_{t+1}=\mathbf{w}_{t}-\eta_t\frac{1}{n}\sum\limits_{i=0}^{n-1}\mathbf{m}_i.
			\end{equation*}
		}
	}
	\caption{Majority-based algorithm to alleviate Byzantine effects.}
	\label{alg:main_algorithm}
\end{algorithm}

\section{Expansion Properties of Bipartite Graphs and Byzantine Resilience}
\label{sec:tanner_expansion_resilience}
The central idea of our work is that concepts from spectral graph theory \cite{chung_spectral} can guide us in the design of the bipartite graph $\bfG$ that specifies the worker-file assignment. In particular, we demonstrate that the spectral properties of the biadjacency matrix of $\bfG$ are closely related to the underlying expansion of the graph and can in turn be related to the scheme's robustness under the omniscient attack model.

A graph with good expansion is one where any small enough set of nodes is guaranteed to have a large number of neighbors. This implies that the Byzantine nodes together process a large number of files which in turn means that their ability to corrupt the majority on many of them is limited. 

%In this section we focus on some properties of bipartite graphs from spectral graph theory and how they can be interpreted in our problem; we discuss their expansion which is a useful metric of resilience of distributed learning schemes to Byzantine nodes. Let the \emph{biadjacency matrix} of the graph (as introduced in Section \ref{sec:formulation}) take the following form.

%AR: An attack in this case is equivalent to picking $q$ vertices on the left (workers) such that if we consider only the edges attached to them, the maximal number of vertices on the right (files) have at least $r'$ of those edges attached to them.

For a bipartite graph $\bfG = (V_1 \cup V_2, E)$ with $|V_1| = K, |V_2| = f$, its biadjacency matrix $H$ is defined as
\begin{equation}
	\label{eq:biadjacency_mat}
	H_{i,j} = \left\{
	\begin{array}{ll}
		1 & \text{ if } (i,j) \in E,\\
		0 & \text{otherwise.}\\
	\end{array} 
	\right.
\end{equation}
We only consider biregular $\bfG$ where the left and right degrees of all nodes are the same; these are denoted $d_L$ and $d_R$ respectively. Let $A = \frac{1}{\sqrt{d_L d_R}}H$ denote the normalized biadjacency matrix. It is known that the eigenvalues of $AA^T$ (or $A^TA$) are $1 = \mu_0 > \mu_1 \geq \cdots \geq \mu_{K-1} \geq \mu_K = \cdots = \mu_{f-1} = 0$ if $K < f$ \cite{chung_spectral}. %(\emph{cf.} \cite{tanner_expansion}). %The second eigenvalue $\mu_1$ is important in quantifying the expansion of $\bfG$.

%$H$ is a $K \times f$ zero-one matrix. The support of each row has size $l$ and the support of each column has size $r$. It has been shown in \cite{ramanujan_murty} that the largest singular value of the biadjacency matrix of any $(d_L,d_R)$-biregular bipartite graph is $\sqrt{d_Ld_R}$ which is equal to the square root of the largest eigenvalue of $H^TH$. Let us normalize $H$ and define
%\begin{equation}
%	\label{eq:normalized_biadjacency_mat}
%	A = [a_{ij}] = \left[\frac{h_{ij}}{\sqrt{d_Ld_R}}\right]
%\end{equation}
%such that the eigenvalues of $M \coloneqq AA^T$ (or $A^TA$) are $1 = \mu_0 > \mu_1 \geq \cdots \geq \mu_{i-1} \geq \mu_i = \cdots = \mu_{f-1} = 0$ if $K < f$ (\emph{cf.} \cite{tanner_expansion}).\footnote{The regime $K<f$ is used here since the number of workers is typically lower than the number of files.} 
%\aditya{I was using script N for the neighborhood previously}
For a subset of the vertices $S$, let $\mathrm{vol}(S)$ denote the number of edges that connect a vertex in $S$ with another vertex not in $S$.
%Then, (i) $N(S)$ denotes its \emph{outer boundary} (or \emph{neighborhood}), i.e., the set of vertices that $S$ is connected to, and (ii) $\mathrm{vol}(S)$ denotes the number of edges that connect a vertex in $S$ with another vertex not in $S$. 
We will use the following lemma \cite{tanner_expansion}. %is proved in .
\begin{lemma}
	\label{lemma:tanner_vol_expansion}
	For any bipartite graph $\bfG=(\calU \cup \calF, \calE)$ and a subset $S$ of $\calU$ (or $\calF$)
	\begin{equation*}
		\frac{\mathrm{vol}(\calN(S))}{\mathrm{vol}(S)} \geq \frac{1}{\mu_1 + (1-\mu_1)\frac{\mathrm{vol}(S)}{|\calE|}}.
	\end{equation*}
\end{lemma}

Now suppose that the bipartite graph $\bfG$ chosen for the worker-file assignment has a second-eigenvalue = $\mu_1$. Based on Lemma \ref{lemma:tanner_vol_expansion}, if $S$ is a set of $q$ Byzantine workers, then $\mathrm{vol}(S) = ql$ since $\bfG$ is bipartite. By substitution, we compute that the number of files collectively being processed by the workers in $S$ is lower bounded as follows.
%\begin{observation}
%	\label{claim:tanner_vol_expansion_workers_files}
%	The cardinality of the outer boundary of the set of $q$ workers $S$ is lower bounded as 
	\begin{equation}
		\label{eq:tanner_vol_expansion_workers_files}
		|\calN(S)| \geq \frac{ql/r}{\mu_1 + (1-\mu_1)\frac{q}{K}} \defeq \beta.
	\end{equation}

\begin{claim}
	\label{claim:max_distortion_matching_attack}
	Let $c_{\mathrm{max}}^{(q)}$ be the maximum number of files one can distort with $q$ Byzantines (since there is a bijection between gradients and files distorting a file is equivalent to distorting the gradient of the samples it contains hence the two terms will be used interchangeably). Then,
	%An optimal attack when a set $S$ of $q$ Byzantines is chosen is when all distorted files in the boundary of $S$ have exactly the minimum necessary edges from $S$, i.e., $r'$ and the remaining boundary (of files) has exactly one edge each from some vertex in $S$. In this case we have shown that (\emph{cf.} Appendix)
	\begin{equation*}
		c_{\mathrm{max}}^{(q)} \leq \frac{l|S| - \beta}{r'-1} = \frac{ql-\beta}{(r-1)/2} \defeq \gamma.
	\end{equation*}
\end{claim}

%The proof is in the Appendix. 
An interesting observation stemming from the above upper bound is that as $\beta$ increases (i.e., as a particular fixed-sized subset of workers collectively processes more files) the maximum number of file gradients they can distort is reduced.

%\begin{claim}
%	\label{claim:max_distortion_matching_attack}
%	An optimal attack when a set $S$ of $q$ Byzantines is chosen is when all distorted files in the boundary of $S$ have exactly the minimum necessary edges from $S$, i.e., $r'$ and the remaining boundary (of files) has exactly one edge each from some vertex in $S$. In this case we have shown that (\emph{cf.} Appendix)
%	\begin{equation*}
%		c_{\mathrm{max}}^{(q)} \leq \frac{l|S| - \beta}{r'-1} = \frac{ql-\beta}{(r-1)/2} \coloneqq \gamma.
%	\end{equation*}
%\end{claim}

%\begin{claim}
%	\label{claim:max_distortion_vanilla}
%	Based on the bipartite graph setup, the volume of the Byzantine vertices in $\calU$ is $lq$. Hence to distort $c$ files we require $ql \geq cr'$ giving rise to the following bound
%	\begin{equation*}
%		c_{\mathrm{max}}^{(q)} \leq \frac{ql}{r'} \coloneqq \delta.
%	\end{equation*}
%\end{claim}
%
%The following claim lower bounds our metric.
%
%\begin{claim}
%	The volume of the corrupted files (assume there are $c$ of them in total) should be at least equal to the number of Byzantines, i.e., $cr' \geq q$ which yields the following lower bound
%	\begin{equation*}
%		c_{\mathrm{max}}^{(q)} \geq \frac{q}{r'}.
%	\end{equation*}
%\end{claim}
%\aditya{Should we combine both constructions into one section?}

\section{Task Assignment}
In this section, we propose two different techniques of constructing the graph $\bfG$ which determines the allocation of gradient tasks to workers in ByzShield. Our graphs have $|\calU| \leq |\calF|$ (i.e., fewer workers than files) and possess optimal expansion properties in terms of their spectrum. %We are interested in graphs such that the number of workers is less than the number of files, i.e., $|\calU| \leq |\calF|$, that possess optimal expansion properties and analyze their performance in adversarial omniscient setups.
 
\subsection{Task Assignment Based on Latin Squares}
\label{sec:MOLS}
We extensively use combinatorial structures known as \emph{Latin squares} \cite{vanlint_wilson_2001} (in short, LS) in this protocol. These designs have been used extensively in error-correcting codes \cite{milenkovic_laendner_2004}. The basic definitions and known results about them that we will need in our exposition are presented below.

%%%%%%%%%%%%%%%%%%%%%%%%%%%%%%%%%%%%%%%%%%%%%%%%%%%%%%%%%%%%%%%%%%%%%%%%%%%%%%%%%%%%%%%%%%%%%%%%%%%%%%%%%%%%%%%%%%%%%%%%%%%%%%%%%%%%%%
% MOLS OF DEGREE 5: METHOD 1: ALIGNED AS ONE ROW
%%%%%%%%%%%%%%%%%%%%%%%%%%%%%%%%%%%%%%%%%%%%%%%%%%%%%%%%%%%%%%%%%%%%%%%%%%%%%%%%%%%%%%%%%%%%%%%%%%%%%%%%%%%%%%%%%%%%%%%%%%%%%%%%%%%%%%
\begin{table}
	\large
	%	\newcolumntype{D}{>{\centering\arraybackslash}m{0.1cm}}
	\newcommand\Ksubtablewidth{0.29\linewidth}
	\newcommand\Kresizetabular{0.9\columnwidth}
	\centering
	\captionsetup[subtable]{position = below, labelformat=empty}
	\caption{A set of three MOLS of degree $5$.}
	\begin{subtable}{\Ksubtablewidth}
		\centering
		{
			\setlength{\extrarowheight}{4pt}
			\resizebox{\Kresizetabular}{!}{
				\begin{tabular}{|P{0.15cm}|P{0.15cm}|P{0.15cm}|P{0.15cm}|P{0.15cm}|}
					\hline
					0&1&2&3&4\\ \hline
					1&2&3&4&0\\ \hline
					2&3&4&0&1\\ \hline
					3&4&0&1&2\\ \hline
					4&0&1&2&3\\ \hline
				\end{tabular}
			}
		}
		\caption{$L_1$}
	\end{subtable}
	\hspace{0cm}
	\begin{subtable}{\Ksubtablewidth}
		\centering
		{
			\setlength{\extrarowheight}{4pt}
			\resizebox{\Kresizetabular}{!}{
				\begin{tabular}{|P{0.15cm}|P{0.15cm}|P{0.15cm}|P{0.15cm}|P{0.15cm}|}
					\hline
					0&1&2&3&4\\ \hline
					2&3&4&0&1\\ \hline
					4&0&1&2&3\\ \hline
					1&2&3&4&0\\ \hline
					3&4&0&1&2\\ \hline
				\end{tabular}
			}
		}
		\caption{$L_2$}
	\end{subtable}
	\hspace{0cm}
	\begin{subtable}{\Ksubtablewidth}
		\centering
		{
			\setlength{\extrarowheight}{4pt}
			\resizebox{\Kresizetabular}{!}{
				\begin{tabular}{|P{0.15cm}|P{0.15cm}|P{0.15cm}|P{0.15cm}|P{0.15cm}|}
					\hline
					0&1&2&3&4\\ \hline
					3&4&0&1&2\\ \hline
					1&2&3&4&0\\ \hline
					4&0&1&2&3\\ \hline
					2&3&4&0&1\\ \hline
				\end{tabular}
			}
		}
		\caption{$L_3$}
	\end{subtable}
%	\hspace{1cm}
%	\begin{subtable}{\Ksubtablewidth}
%		\centering
%		{
%			\setlength{\extrarowheight}{4pt}
%			\resizebox{\Kresizetabular}{!}{
%				\begin{tabular}{|P{0.15cm}|P{0.15cm}|P{0.15cm}|P{0.15cm}|P{0.15cm}|}
%					\hline
%					0&1&2&3&4\\ \hline
%					4&0&1&2&3\\ \hline
%					3&4&0&1&2\\ \hline
%					2&3&4&0&1\\ \hline
%					1&2&3&4&0\\ \hline
%				\end{tabular}
%			}
%		}
%		\caption{$L_4$}
%	\end{subtable}
	\label{table:three_MOLS_degree_5}
	\vspace{-0.1in}
\end{table}
%%%%%%%%%%%%%%%%%%%%%%%%%%%%%%%%%%%%%%%%%%%%%%%%%%%%%%%%%%%%%%%%%%%%%%%%%%%%%%%%%%%%%%%%%%%%%%%%%%%%%%%%%%%%%%%%%%%%%%%%%%%%%%%%%%%%%%

\subsubsection{Basic Definitions and Properties}
\label{sec:ls_theory}
We begin with the definition of a Latin square.

\begin{definition}
	\label{def:latin_square}
	A \emph{Latin square of degree} $l$ is a quadruple $(R,C,S;L)$ where $R, C$ and $S$ are sets with $l$ elements each and $L$ is a mapping $L:R\times C\rightarrow S$ such that for any $i\in R$ and $x\in S$, there is a unique $j\in C$ such that $L(i,j)=x$. Also, any two of $i\in R$, $j\in C$ and $x\in S$ uniquely determine the third so that $L(i,j)=x$.
\end{definition}

%\aditya{use a LS not an LS. Scan the document for more instances}
We refer to the sets $R$, $C$ and $S$ as \emph{rows}, \emph{columns} and \emph{symbols} of the Latin square, respectively. The representation of a LS is an $l \times l$ array $L$ where the cell $L_{ij}$ contains $L(i,j)$. %Definition \ref{def:latin_square} implies that every row of $L$ is a permutation of $S$ and every column of $L$ is also a permutation of $S$. 
%Simple LS of degree $5$ based on the Cayley table of the additive group with $R=C=S=\mathbb{F}_5$ and $L_\alpha(i,j)=\alpha i+j\ (\mathrm{mod}\ 5)$, $\alpha=1,2,3$, is in Table \ref{table:three_MOLS_degree_5}.

%\begin{table*}[t]
%    \centering
%    \caption{A Latin square of degree 3.}
%    \resizebox{0.1\columnwidth}{!}{
%        \begin{tabular}{|P{0.15cm}|P{0.15cm}|P{0.15cm}|}
%        \hline
%        0&1&2\\ \hline
%        1&2&0\\ \hline
%        2&0&1\\ \hline
%        \end{tabular}
%    }\\
%    \label{table:ls_primer_example}
%    \vspace{0.2in}
%\end{table*}

\begin{definition}
	\label{def:orthogonal_ls}
	Two Latin squares $L_1:R\times C\rightarrow S$ and $L_2:R\times C\rightarrow T$ (with the same row and column sets) are said to be \emph{orthogonal} when for each ordered pair $(s,t) \in S \times T$, there is a unique cell $(i,j)\in R \times C$ so that $L_1(i,j)=s$ and $L_2(i,j)=t$.
%	$$L_1(i,j)=s \text{ and } L_2(i,j)=t.$$
\end{definition}
%The orthogonality of $L_1$ and $L_2$ is evident if we superpose\footnote{To be precise, the cell $(i,j)$ of the superposition will be the pair $(L_1(i,j),L_2(i,j))$.} the squares such that each element of $S \times T$ appears exactly once.
A set of $k$ Latin squares of degree $l$ which are pairwise orthogonal are called \emph{mutually orthogonal} or MOLS. It can be shown \cite{vanlint_wilson_2001} that the maximal size of an orthogonal set of Latin squares of degree $l$ is $l-1$.

We use a standard construction which yields $l-1$ MOLS of degree $l$. Initially, pick a prime power $l$. $(i)$ All row, column and symbol sets are to be the elements of $\mathbb{F}_l$ (the finite field of size $l$), i.e., $R=C=S=\mathbb{F}_l=\{0,1,\dots,l-1\}$. $(ii)$ Then, for each nonzero element $\alpha$ in $\mathbb{F}_l$, define $L_\alpha(i,j) := \alpha i+j$ (addition is over $\mathbb{F}_l$). % \aditya{When l is a prime power and not a prime you can do this by modulo -l} 
The $l-1$ squares created in this manner are MOLS since linear equations of the form $ai+bj=s$, $ci+dj=t$ have unique solutions $(i,j)$ provided that $ad-bc \neq 0$ which is the case here.

Table \ref{table:three_MOLS_degree_5} shows an example of the above procedure that yields four MOLS of degree $5$ (only the first three are shown and will be used in the sequel) for $R=C=S=\mathbb{F}_5=\{0,1,\dots,4\}$ and $L_\alpha(i,j)=\alpha i+j\ (\mathrm{mod}\ 5)$, $\alpha=1,2,3$.

%\begin{example}
%\label{ex:two_MOLS_example}
%Let $R=C=S=\mathbb{F}_3=\{0,1,2\}$. Then, we can construct the set of MOLS presented in Table \ref{table:two_MOLS_degree_3}.
%\end{example}

%\begin{table*}
%    \centering
%    \captionsetup[subtable]{position = below}
%    \caption{A pair of MOLS of degree 3.}
%    \begin{subtable}[b]{0.3\linewidth}
%       \centering
%       {
%       \setlength{\extrarowheight}{4pt}
%       \begin{tabular}{|P{0.15cm}|P{0.15cm}|P{0.15cm}|}
%        \hline
%        0&1&2\\ \hline
%        1&2&0\\ \hline
%        2&0&1\\ \hline
%        \end{tabular}
%        }
%       \caption{$L_1=i+j$ (mod $3$)}
%   	\label{table:Cayley_LS_example}
%    \end{subtable}%
%    \quad
%    \begin{subtable}[b]{0.3\linewidth}
%       \centering
%       {
%       \setlength{\extrarowheight}{4pt}
%       \begin{tabular}{|P{0.15cm}|P{0.15cm}|P{0.15cm}|}
%        \hline
%        0&1&2\\ \hline
%        2&0&1\\ \hline
%        1&2&0\\ \hline
%        \end{tabular}
%        }
%       \caption{$L_2=2i+j$ (mod $3$)}
%    \end{subtable}
%    \label{table:two_MOLS_degree_3}
%    \vspace{0.2in}
%\end{table*}

%\aditya{In Algorithm 2, the output is not specified}
%\DecMargin{1em}
\begin{algorithm}[!t]
	%\LinesNotNumbered
	%\SetAlgoNoLine
	\KwIn{Batch size $b$, computation load $l$, redundancy $r\leq l-1$, empty graph $\bfG$ with worker vertices $\calU$ and file vertices $\calF_t$.}
	{
		\abovedisplayskip=0pt
		\belowdisplayskip=0pt
		The PS partitions $B_t$ into $l^2$ disjoint files of $b/l^2$ samples each $$B_t=\left\{B_{t,i}: i=0,1,\dots,l^2-1\right\} = \calF_t.$$\\
		The PS constructs a set of $r$ MOLS $L_1,L_2,\dots,L_{r}$ of degree $l$ with symbols $S=\{0,1,\dots,l-1\}$.\label{alg_step:MOLS}\\
		\For{$k = 0$ to $r-1$}{
			\label{alg_step:oneLS}
			\For{$s = 0$ to $l-1$}{
				PS identifies all cells $(i,j)$ with symbol $s$ in $L_{k+1}$ and connects vertex (file) $B_{t,il+j}$ to vertex (worker) $U_{kl+s}$ in $\bfG$.
			}
		}
		
		\label{alg_step:r3_end}
	}
	\caption{Proposed MOLS-based assignment.}
	\label{alg:MOLS_placement}
\end{algorithm}

\subsubsection{Redundant Task Allocation}
\label{sec:allocation}
%To allocate the batch of an iteration, $B_t$, to workers, first, we will partition $B_t$ into $f=l^2$ disjoint \emph{files} $B_{t,0},B_{t,2},\dots,B_{t,l^2-1}$ of $b/l^2$ samples each where $l$ is the degree of a set of MOLS constructed as in Sec. \ref{sec:ls_theory}. 
%We will arrange the files on a $l\times l$ grid $L$ where $L_{ij}$ is populated with the file $B_{t,il+j}$, where $i,j\in\{0,1,\dots,l-1\}$. Each of the files will be assigned repetitively to $r$ workers, where $1\leq r \leq l-1$. Let us use $K$ workers where $K=rl$ and think of the placement as its corresponding bipartite graph $\bfG = (\calU \cup \calF_t, \calE)$, where $\calU =  \{U_0,U_1\dots,U_{K-1}\}$ and $\calF_t = \left\{B_{t,i}: i=0,1,\dots,l^2-1\right\}$. The exact placement policy is illustrated in Algorithm \ref{alg:MOLS_placement}. 

%AR:reworded to save space
To allocate the batch of an iteration, $B_t$, to workers, first, we will partition $B_t$ into $f=l^2$ disjoint files $B_{t,0},B_{t,2},\dots,B_{t,l^2-1}$ of $b/l^2$ samples each where $l$ is the degree of a set of MOLS constructed as in Sec. \ref{sec:ls_theory}. Following this we construct a bipartite graph specifying the placement according to the MOLS (see Algorithm \ref{alg:MOLS_placement}).

The following example showcases the proposed protocol of Algorithm \ref{alg:MOLS_placement} using the MOLS from Table \ref{table:three_MOLS_degree_5}.

\begin{example}
	\label{ex:ex2_mols_25}
	Consider $K=15$ workers $U_0,\dots,U_{14}$, $l=5$ and $r=3$. Based on our protocol the $l^2=25$ files of each batch $B_t$ would be arranged on a grid $L$, as defined above.
%	\begin{equation*}
%		L =
%		{
%			\setlength{\extrarowheight}{1pt}
%			\begin{tabular}{|P{0.7cm}|P{0.7cm}|P{0.7cm}|P{0.7cm}|P{0.7cm}|}
%				\hline
%				$B_{t,0}$&$B_{t,1}$&$B_{t,2}$&$B_{t,3}$&$B_{t,4}$\\ \hline
%				$B_{t,5}$&$B_{t,6}$&$B_{t,7}$&$B_{t,8}$&$B_{t,9}$\\ \hline
%				$B_{t,10}$&$B_{t,11}$&$B_{t,12}$&$B_{t,13}$&$B_{t,14}$\\ \hline
%				$B_{t,15}$&$B_{t,16}$&$B_{t,17}$&$B_{t,18}$&$B_{t,19}$\\ \hline
%				$B_{t,20}$&$B_{t,21}$&$B_{t,22}$&$B_{t,23}$&$B_{t,24}$\\ \hline
%			\end{tabular}
%		}
%	\end{equation*}
	Since $r=3$, the construction (see line \ref{alg_step:MOLS} of the algorithm) involves the use of $3$ MOLS of degree $l=5$, shown in Table \ref{table:three_MOLS_degree_5}. For the LS $L_1$ ($k=0$ in step \ref{alg_step:oneLS} of the algorithm) choose, e.g., symbol $s=0$. The locations of $s=0$ in $L_1$ are $(0,0),(1,4),(2,3),(3,2) \text{ and }(4,1)$ and the files in those cells in $L$ are $B_{t,0}, B_{t,9}, B_{t,13}, B_{t,17} \text{ and } B_{t,21}.$
	Hence, $U_{0}$ will receive those files from the PS. The complete file assignment for the cluster is shown in Table \ref{table:ex2_mols_25_file_assignment} (the index $i$ of file $B_{t,i}$ has been used instead for brevity).
	
	%Based on steps \ref{alg_step:r1_start}-\ref{alg_step:r1_end} of Algorithm \ref{alg:MOLS_placement}, we will initially use workers $\{U_0,U_1,\dots,U_4\}$ and assign a row of $L$ to each of them. For example, $U_0$ will be handling files $\{B_{t,i}: i=0,1,\dots,4\}$. Next, we proceed to steps \ref{alg_step:r2_start}-\ref{alg_step:r2_end} in which $\{U_5,\dots, U_{9}\}$ receive columns $\{0,1,\dots,4\}$ of $L$, respectively. Since $r=5$, the last step (see lines \ref{alg_step:r3_start}-\ref{alg_step:r3_end} of the algorithm) involves the use of 3 MOLS of degree $l=5$, which are shown in Table \ref{table:three_MOLS_degree_5} (see Section \ref{sec:ls_theory} for the method used). We can now instantiate the remaining workers $U_{10},\dots,U_{24}$. For the first LS $L_1$ ($k=1$ in step \ref{alg_step:r3_oneLS} of the algorithm) choose, for example, symbol $s=0$. The locations of $s=0$ in $L_1$ are
	%$$(0,0),(1,4),(2,3),(3,2) \text{ and }(4,1)$$
	%and the files in those cells in $L$ are
	%$$B_{t,0}, B_{t,9}, B_{t,13}, B_{t,17} \text{ and } B_{t,21}.$$
	%Hence, $U_{10}$ will receive those files from the PS. The file assignment for the cluster is shown in Table \ref{table:ex2_mols_25_file_assignment} (the index $i$ of file $B_{t,i}$ has been used instead for brevity).
\end{example}

It is evident that each worker stores $l$ files. Another observation that follows from the placement policy (\emph{cf.} Algorithm \ref{alg:MOLS_placement}) and Definition \ref{def:latin_square} is that any two workers populated based on the same Latin square do not share any files while Definition \ref{def:orthogonal_ls} implies that a pair of workers populated based on two orthogonal Latin squares should share exactly one file.

%; we extend this argument to a more general one that we will utilize to prove our robustness (\emph{cf.} Lemma \ref{lemma:MOLS_common_file}). 
%First, any node among those assigned to files row-wise from $L$ trivially has one-file in common with any node assigned column-wise. The same holds between any node assigned based on a symbol $s$ of any of the MOLS and any row node. This is by the definition of a LS (\emph{cf.} Definition \ref{def:latin_square}) since for a particular symbol $s$ there cannot be more than one occurrences of $s$ on the same row of a LS; an identical result holds for the intersection of column nodes and Latin squares. Last, 

%\aditya{Overall comment: A head to head comparison with prior work in terms of the theoretical guarantees is lacking}
\subsection{Task Assignment Based on Direct Graph Constructions}
\label{sec:bipartite_constructions}
We next focus on direct construction of bipartite graphs for the task assignment. In this regard, we investigate Ramanujan bigraphs which have been shown to be optimal expanders \cite{alon_1986}. 
% The graph structure is encoded in its adjacency matrix which will be used to assign tasks to workers.
A formal definition follows.

\begin{definition}
	A $(d_L,d_R)$-biregular bipartite graph is a \emph{Ramanujan} bigraph if the second largest singular value of its biadjacency matrix $H$ (from Eq. \eqref{eq:biadjacency_mat}) is less than or equal to $\sqrt{d_L - 1} + \sqrt{d_R - 1}$.
\end{definition}

Equivalently, one can express the property in terms of the second largest eigenvalue of $AA^T$, as defined in Section \ref{sec:tanner_expansion_resilience}. 
%(see also  \cite{optimal_ramanujan_feng}). 
This equivalence will  be discussed in Section \ref{sec:distortion_analysis} and in the corresponding proofs in the Appendix. Overall, we seek Ramanujan bigraphs which provably achieve the smallest possible value of the second largest eigenvalue of $AA^T$.

\begin{table}
	\newcommand\Ksubtablewidth{0.3\linewidth}
	\centering
	\captionsetup[subtable]{position = below}
	\caption{File allocation for $l=5$, $r=3$ based on MOLS.}
	\begin{subtable}{\Ksubtablewidth}
		\centering
		{
			\setlength\tabcolsep{1.5pt}
			\setlength{\extrarowheight}{2pt}
			\resizebox{\columnwidth}{!}{
				\begin{tabular}{|P{0.8cm}|P{2.2cm}|}
					\hline
					Node&Stores\\
					\hline
					$U_{0}$&$0,9,13,17,21$\\ \hline
					$U_{1}$&$1,5,14,18,22$\\ \hline
					$U_{2}$&$2,6,10,19,23$\\ \hline
					$U_{3}$&$3,7,11,15,24$\\ \hline
					$U_{4}$&$4,8,12,16,20$\\ \hline
				\end{tabular}
			}
		}
		\caption{1st replica.}
	\end{subtable}
	\hspace{0cm}
	\begin{subtable}{\Ksubtablewidth}
		\centering
		{
			\setlength\tabcolsep{1.5pt}
			\setlength{\extrarowheight}{2pt}
			\resizebox{\columnwidth}{!}{
				\begin{tabular}{|P{0.8cm}|P{2.2cm}|}
					\hline
					Node&Stores\\
					\hline
					$U_{5}$&$0,8,11,19,22$\\ \hline
					$U_{6}$&$1,9,12,15,23$\\ \hline
					$U_{7}$&$2,5,13,16,24$\\ \hline
					$U_{8}$&$3,6,14,17,20$\\ \hline
					$U_{9}$&$4,7,10,18,21$\\ \hline
				\end{tabular}
			}
		}
		\caption{2nd replica.}
	\end{subtable}
	\hspace{0cm}
	\begin{subtable}{\Ksubtablewidth}
		\centering
		{
			\setlength\tabcolsep{1.5pt}
			\setlength{\extrarowheight}{2pt}
			\resizebox{\columnwidth}{!}{
				\begin{tabular}{|P{0.8cm}|P{2.2cm}|}
					\hline
					Node&Stores\\
					\hline
					$U_{10}$&$0,7,14,16,23$\\ \hline
					$U_{11}$&$1,8,10,17,24$\\ \hline
					$U_{12}$&$2,9,11,18,20$\\ \hline
					$U_{13}$&$3,5,12,19,21$\\ \hline
					$U_{14}$&$4,6,13,15,22$\\ \hline
				\end{tabular}
			}
		}
		\caption{3rd replica.}
	\end{subtable}
	%	\hspace{0cm}
	%	\begin{subtable}{\Ksubtablewidth}
	%		\centering
	%		{
	%		\setlength\tabcolsep{1.5pt}
	%		\setlength{\extrarowheight}{2pt}
	%		\resizebox{\columnwidth}{!}{
	%			\begin{tabular}{|P{0.8cm}|P{2.2cm}|}
	%				\hline
	%				Node&Stores\\
	%				\hline
	%				$U_{15}$&$0,6,12,18,24$\\ \hline
	%				$U_{16}$&$1,7,13,19,20$\\ \hline
	%				$U_{17}$&$2,8,14,15,21$\\ \hline
	%				$U_{18}$&$3,9,10,16,22$\\ \hline
	%				$U_{19}$&$4,5,11,17,23$\\ \hline
	%			\end{tabular}
	%		}
	%	}
	%	\caption{4th replica.}
	%	\end{subtable}
	\label{table:ex2_mols_25_file_assignment}
	\vspace{-0.1in}
\end{table}

\subsubsection{Redundant Task Allocation}
\label{sec:ramanujan}
There are many ways to construct a $(d_L,d_R)$-biregular bipartite graph with the Ramanujan property \cite{eigen_exp_bipar_hoholdt, sipser_spielman_1996}. We use the one presented in \cite{ramanujan_bigraphs_Shantanu}. This method yields a $(s,m)$ or an $(m,s)$ biregular graph (depending on the relation between $m$ and $s$) for $m\geq2$ and a prime $s$. It is based on LDPC ``array code" matrices \cite{ldpc} and proceeds as follows.
\begin{itemize}
	\item \textbf{Step 1}: Pick integer $m \geq 2$ and prime $s$.
	\item \textbf{Step 2}: Define a cyclic shift permutation matrix $P_{s\times s}$
	\begin{equation*}
		P_{ij} = \left\{
		\begin{array}{ll}
			1 & \text{ if } j \equiv i-1\ (\mathrm{mod}\ s),\\
			0 & \text{otherwise} \\
		\end{array} 
		\right.
	\end{equation*}
	where $i=1,2,\dots,s$ and $j=1,2,\dots,s$.
	\item \textbf{Step 3}: Construct matrix $B$ as
	\begin{equation*}
		B=
		\begin{bmatrix}
			I_{s} & I_{s} & I_{s} & \cdots & I_{s}\\
			I_{s} & P & P^2 & \cdots & P^{m-1}\\
			I_{s} & P^2 & P^4 & \cdots & P^{2(m-1)}\\
			I_{s} & P^3 & P^6 & \cdots & P^{3(m-1)}\\
			\vdots & \vdots & \vdots & \vdots & \vdots\\
			I_{s} & P^{s-1} & P^{2(s-1)} & \cdots & P^{(m-1)(s-1)}\\
		\end{bmatrix}.
	\end{equation*}
	$B$ is a $s^2 \times ms$ matrix constructed as a $s \times m$ block matrix. If $m \geq s$, then $H=B$ is chosen to be the biadjacency matrix of the bipartite graph, otherwise $H=B^T$ is chosen. In both cases, $H$ represents a Ramanujan bigraph. Note that $H$ is a zero-one biadjacency matrix since $P$ and hence all of its powers are permutation matrices. 
\end{itemize}

As mentioned earlier, we will assume $K \leq f$. The rows of $H$ will therefore correspond to workers and the columns will correspond to files in both of the above cases. Since all blocks of $H$ are permutation matrices the support of each row (size of storage of worker) and column (replication of file) can be easily derived. This yields two cases when this construction is applied to our setup
\begin{equation}
	\label{eq:ramanujan_cluster_setups}
	(K,f,l,r) = \left\{
	\begin{array}{ll}
		(ms,s^2,s,m) & \text{if } m < s,\\
		(s^2,ms,m,s) & \text{otherwise}.
	\end{array} 
	\right.
\end{equation}

We will refer to the case $m<s$ as \emph{Ramanujan Case 1} and to the other case as \emph{Ramanujan Case 2}. We can have setups with identical parameters between the MOLS formulation and Case 1; specifically, one can choose the same computation load $l$ (for prime $l$) and replication factor $r$ such that $r < l$ which in both schemes will make use of $K=rl$ workers and $f=l^2$ files. The need for a prime $l$ makes the Ramanujan scheme less flexible in terms of choosing the computation load, though. Ramanujan Case 2 requires $r \geq l$ hence we examine this method separately.
%Cases 2 and 3 of Ramanujan schemes lead to identical parameters as defined in Eq. \eqref{eq:ramanujan_cluster_setups} and direct comparisons of their performance can be made.

In our schemes, we need to allocate $K$ workers such that $K$ factorizes as $K = rl$ or $K = r^2$, requiring a prime $l$, a prime power $l$ or a prime $r$. We argue that these conditions are not restrictive and many such choices of $K$ exist on modern distributed platforms such as Amazon EC2 few of which are presented in this paper.

\section{Distortion Fraction Analysis}
\label{sec:distortion_analysis}
In this section, we perform a worst-case analysis of the fraction of distorted files (defined as $\hat{\epsilon}=c_{\mathrm{max}}^{(q)}/f$) incurred by various aggregation methods, including that of ByzShield. 

A few reasons motivate this analysis. Our deep learning experiments (upcoming Section \ref{sec:experiment_results}) show that $\hat{\epsilon}$ is a very useful indicator of the convergence of a model's training. Also, this comparison shows the superiority of ByzShield in terms of theoretical robustness guarantees since we achieve a much smaller $\hat{\epsilon}$ for the same $q$. Finally, we demonstrate via numerical experiments that the quantity $\gamma$ is a tight upper bound on the exact value of $c_{\mathrm{max}}^{(q)}$ for ByzShield.% We also present exhaustive numerical simulations for these metrics.

We next compute the spectrum of $AA^T$, as defined in Section \ref{sec:tanner_expansion_resilience}, of our utilized constructions. The authors of \cite{ramanujan_bigraphs_Shantanu} [Theorem 6] have computed the set of singular values of $H$ of Ramanujan Cases 1 and 2. In order to have a direct comparison between the spectral properties of the MOLS scheme and these Ramanujan graphs, we have normalized $H$ and computed the eigenvalues of the corresponding $AA^T$ instead. For the MOLS scheme, the normalized biadjacency matrix $A$ has $d_L=l$ and $d_R=r$. We denote an eigenvalue $x$ with algebraic multiplicity $y$ with $(x,y)$. All proofs for this section are in the Appendix.

\begin{lemma}
	\label{lemma:spectrum_of_all_schemes}
	%There are the following cases for the spectrum of $AA^T$ for the Ramanujan bigraphs constructed as in \cite{ramanujan_bigraphs_Shantanu} and the proposed MOLS bigraph.
	\hfill
	\begin{itemize}
		\item \textbf{MOLS and Ramanujan Case 1}: $(AA^T)_{MOLS}$ and $(AA^T)_{Ram. 1}$ require $2<r<l$ and have spectrum
		$$\{(1,1), (1/r,r(l-1)), (0,r-1)\}.$$
		\item \textbf{Ramanujan Case 2}: $(AA^T)_{Ram. 2}$ requires $r\leq l$, $r|l$ and has spectrum
		$$\{(1,1), (1/r,r(r-1)), (0,r-1)\}.$$
	\end{itemize}
\end{lemma}

%\begin{lemma}
%	\label{lemma:ramanujan_shantau_spectrum}
%	There are the following cases for the spectrum of $AA^T$ for the Ramanujan bigraphs constructed as in \cite{ramanujan_bigraphs_Shantanu} [Theorem 6, Cases 1 and 2].
%	\begin{itemize}
%		\item \textbf{Case 1}: If $2 \leq m < s$, then $AA^T$ has spectrum
%		$$\{(1,1), (1/m,m(s-1)), (0,m-1)\}.$$
%		\item \textbf{Case 2}: If $m \geq s$ and $s|m$, then $AA^T$ has spectrum
%		$$\{(1,1), (1/s,(s-1)s), (0,s-1)\}.$$
%		%		\item \textbf{Case 3}: If $m \geq s$ and $s \nmid m$, let $k=m \text{ mod } s$, then $AA^T$ has spectrum with algebraic multiplicities
%		%		$$\left\{(1,1), \left(\frac{m+s-k}{sm},(s-1)k\right), \left(\frac{m-k}{sm}, (s-1)(s-k)\right), (0,s-1)\right\}$$
%	\end{itemize}
%\end{lemma}

The eigenvalues are sorted in decreasing order of magnitude. Interestingly, $(AA^T)_{Ram. 1}$ has exactly the same spectrum as if the graph was constructed from a set of MOLS. Overall, the parameters $l,r$, number of workers $K=rl$, number of files $f=l^2$ and the second largest eigenvalue of $AA^T$, $\mu_1=1/r$, are all identical across the two schemes.

\subsection{Upper Bounds of $\hat{\epsilon}^{ByzShield}$ Based on Graph Expansion}
\label{sec:epsilon_upper_bounds}
%We will now use the spectral properties of the underlying graphs of ByzShield and their adjacency matrices in order to provide worst-case upper bounds on $\hat{\epsilon}$ for any value of $q$. This attack analysis relies on Section \ref{sec:tanner_expansion_resilience} (computation of $\beta$ and Lemma \ref{lemma:tanner_vol_expansion}). 
%To that end, we substitute $\mu_1$ (for a particular graph) in Eq. \eqref{eq:tanner_vol_expansion_workers_files}. We then use Lemma \ref{lemma:tanner_vol_expansion} to provide upper bounds on $c_{\mathrm{max}}^{(q)}$ and subsequently on $\hat{\epsilon}=c_{\mathrm{max}}^{(q)}/f$. 
%All proofs are in the Appendix.

\subsubsection{MOLS-based Graphs}
\label{sec:MOLS_spectral}
%Then, for $AA^T$ we have computed the eigenvalues summarized below.
%\begin{lemma}
%	\label{lemma:MOLS_eigenvalues}
%	The spectrum of $AA^T$ is the following
%	$$\{(1,1), (1/r,(l-1)r), (0,r-1)\}.$$
%%	\begin{itemize}
%%		\item $\lambda = 1$ with algebraic multiplicity $1$.
%%		\item $\lambda = 1/r$ with algebraic multiplicity $(l-1)r$.
%%		\item $\lambda = 0$ with algebraic multiplicity $r-1$.
%%	\end{itemize}
%\end{lemma}

Based on Lemma \ref{lemma:spectrum_of_all_schemes}, $\mu_1=1/r$ is the second largest eigenvalue of $(AA^T)_{MOLS}$. Substituting in the expression of Eq. \eqref{eq:tanner_vol_expansion_workers_files}, Claim \ref{claim:max_distortion_matching_attack} gives us the exact value for $\gamma$. It is a tight upper bound on $c_{\mathrm{max}}^{(q)}$ for the MOLS protocol and by normalizing it we conclude that an optimal attack with $q$ Byzantines, distorts a fraction of the files which is at most
\begin{equation*}
	\hat{\epsilon}^{MOLS} \leq \frac{\gamma}{f} = \frac{\frac{2q^2}{rl^2}}{1+(r-1)\frac{q}{rl}}.
\end{equation*}

\subsubsection{Ramanujan Graphs}
 By Lemma \ref{lemma:spectrum_of_all_schemes}, the underlying graph of the MOLS assignment is a Ramanujan graph and all results of Section \ref{sec:MOLS_spectral} carry over verbatim to Case 1. 

Case 2 needs to be treated separately since it leads to a different cluster setup with $K=s^2=r^2$ workers and $f=rl$ files with an underlying graph with $\mu_1=1/s=1/r$. A simple computation yields
\begin{equation*}
	\hat{\epsilon}^{Ram. 2} \leq \frac{\gamma}{f} = \frac{\frac{2q^2}{r^2}}{r + (r-1)\frac{q}{r}}.
\end{equation*}

\subsection{Exact Value of $\hat{\epsilon}^{ByzShield}$ in the Regime $q \leq r$}
\label{sec:q_less_r_exact}
%Section \ref{sec:epsilon_upper_bounds} was devoted to computations of tight upper bounds on $\hat{\epsilon}$ under omniscient attacks on our system.
When $q \leq r$, i.e., in the regime where there are only a few Byzantine workers, we can provide exact values of $\hat{\epsilon}$ for our constructions.
%This section provides exact results on $\hat{\epsilon}$ for the case of $q \leq r$ under optimal Byzantine attacks. 
In particular we pick the $q$ workers to be Byzantine such that the multiset of the files stored collectively across them (also referred to as \emph{multiset sum}) has the maximal possible number of files repeated at least $r'$ times. In this way, the majority of the copies of those files is distorted and the aggregation produces erroneous gradients.

\begin{claim}
	\label{claim:r_less_q_distortions}
	The maximum distortion fraction incurred by an optimal attack for the regime $q\leq r$ is characterized as follows.
	\begin{itemize}
		\item If $r = 3$, then 
		$\hat{\epsilon}^{ByzShield} = 
		\left\{
		\begin{array}{ll}
			0 & \text{if } q<2,\\
			1/f & \text{if } q=2,\\
			3/f & \text{if } q=3.
		\end{array} 
		\right.$
		
%		\begin{equation*}
%			\hat{\epsilon}^{ByzShield} = \left\{
%			\begin{array}{ll}
%				0 & \text{if } q<2,\\
%				1/f & \text{if } q=2,\\
%				3/f & \text{if } q=3.
%			\end{array} 
%			\right.
%		\end{equation*}

%		\item If $r>3$, then $\hat{\epsilon}^{ByzShield}\leq 2/f$.
		\item If $r>3$, then 
		$\hat{\epsilon}^{ByzShield} = 
		\left\{
		\begin{array}{ll}
			0 & \text{if } q<r',\\
			1/f & \text{if } r'\leq q < r,\\
			2/f & \text{if } q=r.
		\end{array} 
		\right.$
		
%		\begin{equation*}
%			\hat{\epsilon}^{ByzShield} = \left\{
%			\begin{array}{ll}
%				0 & \text{if } q<r',\\
%				1/f & \text{if } r'\leq q < r,\\
%				2/f & \text{if } q=r.
%			\end{array} 
%			\right.
%		\end{equation*}

	\end{itemize}
	
\end{claim}

\subsection{Comparisons With Prior Work}
\label{sec:simulations}

%\aditya{The section needs to clearly discuss which methods were used for comps and why?}
%\aditya{Needs to be divided into parts discussing theoretical comp and then experimental comp}
We compare with the values of $\hat{\epsilon}$ achieved by \emph{baseline} and other redundancy-based methods. Baseline approaches do not involve redundancy or majority voting; their aggregation is applied directly to the $K$ gradients returned by the workers and hence $f=K$ and $c_{\mathrm{max}}^{(q)}=q$ which yields a distortion fraction $\hat{\epsilon}=q/K$. We also compare the actual values of $c_{\mathrm{max}}^{(q)}$ for ByzShield against the theoretical upper bounds based on the expansion analysis in Section \ref{sec:epsilon_upper_bounds}.

\subsubsection{Achievable $\hat{\epsilon}$ of Prior Work Under Worst-case Attacks}
\label{sec:simulations_theory}
We now present a direct comparison between our work and state-of-the-art schemes that are redundancy-based. \emph{DETOX} in \cite{detox} and \emph{DRACO} in \cite{draco} both use the same \emph{Fractional Repetition Code} (FRC) for the assignment and both use majority vote aggregation, hence we will treat them in a unified manner. Our goal is to establish that redundancy is not enough to achieve resilience to adversaries; rather, a careful task assignment is of major importance. The basic placement of FRC is splitting the $K$ workers into $K/r$ groups. Within a group, all workers are responsible for the same part of a batch of size $br/K$ and a majority vote aggregation is executed in the group. Their method fundamentally relies on the fact that the Byzantines are chosen at random for each iteration and hence the probability of at least $r'$ Byzantines being in the same majority-vote group is low on expectation. However, in an omniscient setup, if an adversary chooses the Byzantines such that at least $r'$ workers in each group are adversarial, then it is evident that all corresponding gradients will be distorted. Such an attack can distort the votes in $\lfloor \frac{q}{r'} \rfloor$ groups. To formulate a direct comparison with ByzShield let us measure $\hat{\epsilon}$ in this worst-case scenario. For FRC, this is equal to the number of affected groups multiplied by the number of samples per group and normalized by $b$, i.e.,
\begin{equation*}
	\hat{\epsilon}^{FRC} = \frac{\lfloor \frac{q}{r'} \rfloor \times br/K}{b}=\lfloor \frac{q}{r'} \rfloor \times r/K.
\end{equation*}
We pinpoint that at the regime $q\leq r'$ all of the aforementioned schemes, including ours, achieve perfect recovery since there are not enough adversaries to distort any file. In addition, DRACO would fail in the regime $q>r'$ while ByzShield still demonstrates strong robustness results.

%%%%%%%%%%%%%%%%%%%%%%%%%%%%%%%%%%%%%%%%%%%%%%%%%%%%%%%%%%%%%%%%%%%%%%%%%%%%%%%%%%%%%%%%%%%%%%%%%%%%%%%%%%%%%%%%%%%%%%%%%%%%%%%%%%%%%%
% TO BE INCLUDED
%%%%%%%%%%%%%%%%%%%%%%%%%%%%%%%%%%%%%%%%%%%%%%%%%%%%%%%%%%%%%%%%%%%%%%%%%%%%%%%%%%%%%%%%%%%%%%%%%%%%%%%%%%%%%%%%%%%%%%%%%%%%%%%%%%%%%%
%Hence, if it is known that the FRC has been used this choice for the Byzantine set can be carried out in $\mathcal{O}(K)$ time. If an adversary does not have this knowledge but knows the batch assignment it will need to consider all pairs of workers to figure out groups which will need $\mathcal{O}(K^2)$ comparisons. In both cases, this cost is asymptotically negligible.
%%%%%%%%%%%%%%%%%%%%%%%%%%%%%%%%%%%%%%%%%%%%%%%%%%%%%%%%%%%%%%%%%%%%%%%%%%%%%%%%%%%%%%%%%%%%%%%%%%%%%%%%%%%%%%%%%%%%%%%%%%%%%%%%%%%%%%

%In ByzShield, the fraction is equal to $\hat{\epsilon} = c_{\mathrm{max}}^{(q)}/f$.

\begin{table}[!t]
	\centering
	\caption{Distortion fraction evaluation for MOLS-based assignment for $(K,f,l,r)=(15,25,5,3)$ and comparison.}
	\label{table:MOLS_distortion_test_l5r3}
	%\resizebox{0.8\columnwidth}{!}{
	\setlength\tabcolsep{2pt}
	\begin{tabular}{ |P{0.6cm}||P{0.9cm}|P{1.8cm}|P{1.5cm}|P{1.1cm}|P{1.1cm}| }
		\hline
		$q$ & $c_{\mathrm{max}}^{(q)}$ & $\hat{\epsilon}^{ByzShield}$ & $\hat{\epsilon}^{Baseline}$ & $\hat{\epsilon}^{FRC}$ & $\gamma$ \\
		\hline
		$2$ & 1 & 0.04 & 0.13 & 0.2 & 2.11 \\
		$3$ & 3 & 0.12 & 0.2 & 0.2 & 4.29 \\
		$4$ & 5 & 0.2 & 0.27 & 0.4 & 6.96 \\
		$5$ & 8 & 0.32 & 0.33 & 0.4 & 10 \\
		$6$ & 12 & 0.48 & 0.4 & 0.6 & 13.33 \\
		$7$ & 14 & 0.56 & 0.47 & 0.6 & 16.9 \\
		\hline
	\end{tabular}
	%}
	\vspace{-0.1in}
\end{table}

\subsubsection{Numerical Simulations}
For ByzShield, we ran exhaustive simulations, %\footnote{\href{https://tinyurl.com/y2fcb6yp}{https://tinyurl.com/y2fcb6yp}}
i.e, we computed $c_{\mathrm{max}}^{(q)}$ considering all adversarial choices of $q$ out of $K$ workers. The evaluation of the MOLS-based allocation scheme of Section \ref{sec:MOLS} for Example \ref{ex:ex2_mols_25} with $(l,r)=(5,3)$ is in Table \ref{table:MOLS_distortion_test_l5r3} where for different values of $q$ we report the simulated $c_{\mathrm{max}}^{(q)}$, the corresponding $\hat{\epsilon}$ and the upper bound $\gamma$. Since a Ramanujan-based scheme of Case 1 has the same properties, the upper bounds are expected to be identical (however the actual task assignment is in general different). An interesting observation is that the simulations of the actual value of $c_{\mathrm{max}}^{(q)}$ were also identical across the two; thus, both have been summarized in the same Table \ref{table:MOLS_distortion_test_l5r3}. More simulations including Ramanujan Case 2 are shown in Section \ref{supplementary:extra_e_hat_simulations} of the Appendix. From these tables we deduce that $\gamma$ is a very accurate worst-case approximation of  $c_{\mathrm{max}}^{(q)}$.

Comparing with the achieved $\hat{\epsilon}^{FRC}$ computed as in Section \ref{sec:simulations_theory}, note that ByzShield can tolerate a higher number of Byzantines while consistently maintaining a small fraction $\hat{\epsilon}$. In contrast, a worst-case attack on FRC raises this error very quickly with respect to $q$. On average, $\hat{\epsilon}^{ByzShield}=0.64\hat{\epsilon}^{FRC}$ in Table \ref{table:MOLS_distortion_test_l5r3}. Finally, ByzShield has a benefit over baseline methods for up to $q=5$ with respect to $\hat{\epsilon}$.
%\footnote{FRC has the assumptions $r|K$ and $K|b$ which are satisfied for our comparison.}
% and $\hat{\epsilon}^{ByzShield}=0.56\hat{\epsilon}^{DETOX}$ in Table \ref{table:Ramanujan_distortion_test_l5r5}.

%\begin{table*}[!t]
%\centering
%\caption{BLA BLA BLA}
%\resizebox{0.8\columnwidth}{!}{
%\begin{tabular}{ |P{3cm}||P{1.9cm}|P{1.4cm}|P{1.4cm}|P{1.4cm}| }
%\hline
%Number of Byzantines ($q$) & Simulated $c_{\mathrm{max}}^{(q)}$ & $\beta$  & $\gamma$ & $\delta$\\
%\hline
%$3$ & \multicolumn{4}{c|}{\multirow{10}{*}{Identical to Table TABLE}}\\
%$4$ & \multicolumn{4}{c|}{}\\
%$5$ & \multicolumn{4}{c|}{}\\
%$6$ & \multicolumn{4}{c|}{}\\
%$7$ &  \multicolumn{4}{c|}{}\\
%$8$ &  \multicolumn{4}{c|}{}\\
%$9$ &  \multicolumn{4}{c|}{}\\
%$10$ &  \multicolumn{4}{c|}{}\\
%$11$ &  \multicolumn{4}{c|}{}\\
%$12$ &  \multicolumn{4}{c|}{}\\
%\hline
%\end{tabular}
%}
%%\vspace{-0.2in}
%\end{table*}

\begin{figure*}[!htb]
	\minipage{0.32\textwidth}
	\includegraphics[width=\linewidth]{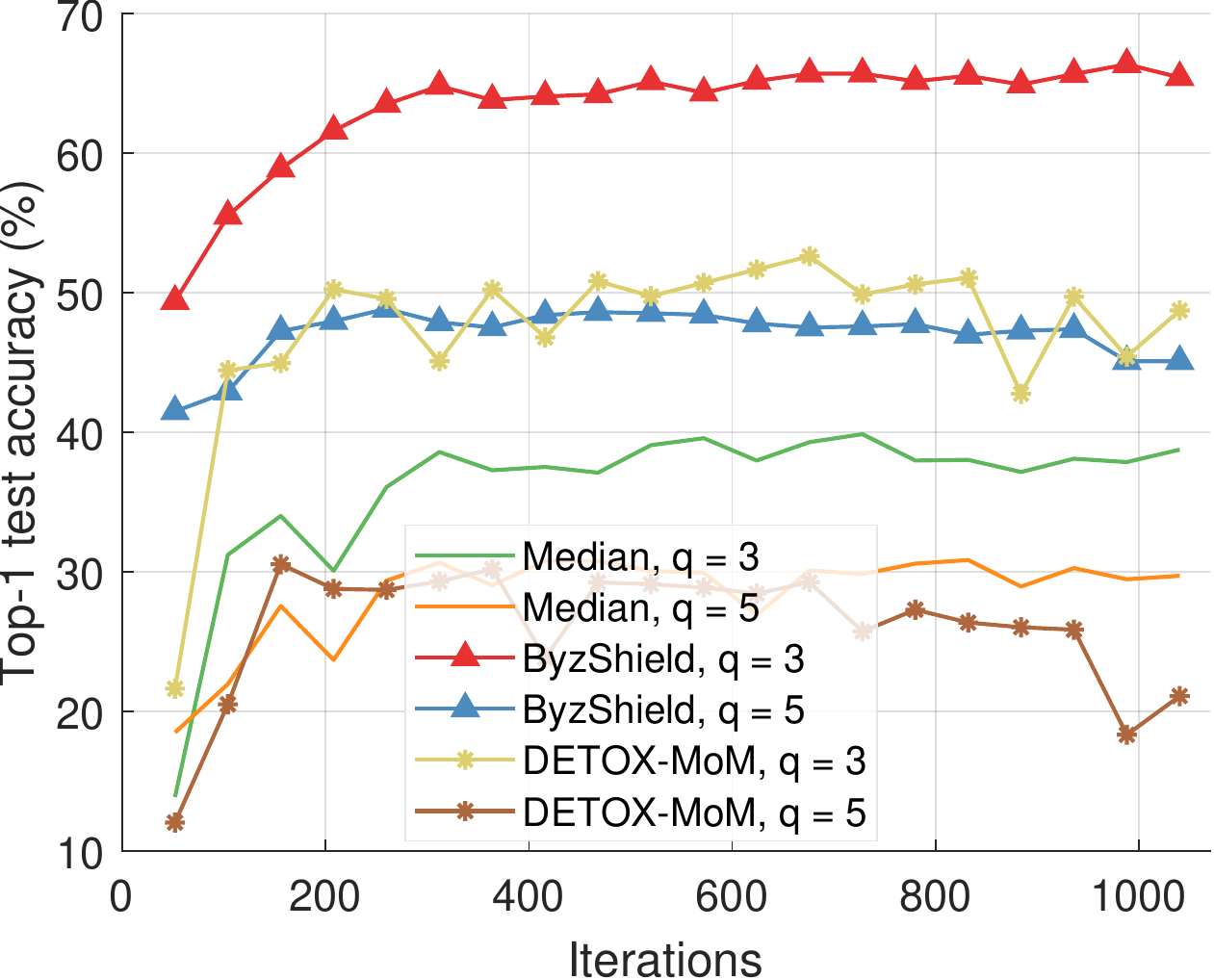}
	\caption{\emph{ALIE} attack and median-based defenses (CIFAR-10).}
	\label{fig:top1_fig_62}
	\endminipage\hfill
	\minipage{0.32\textwidth}
	\includegraphics[width=\linewidth]{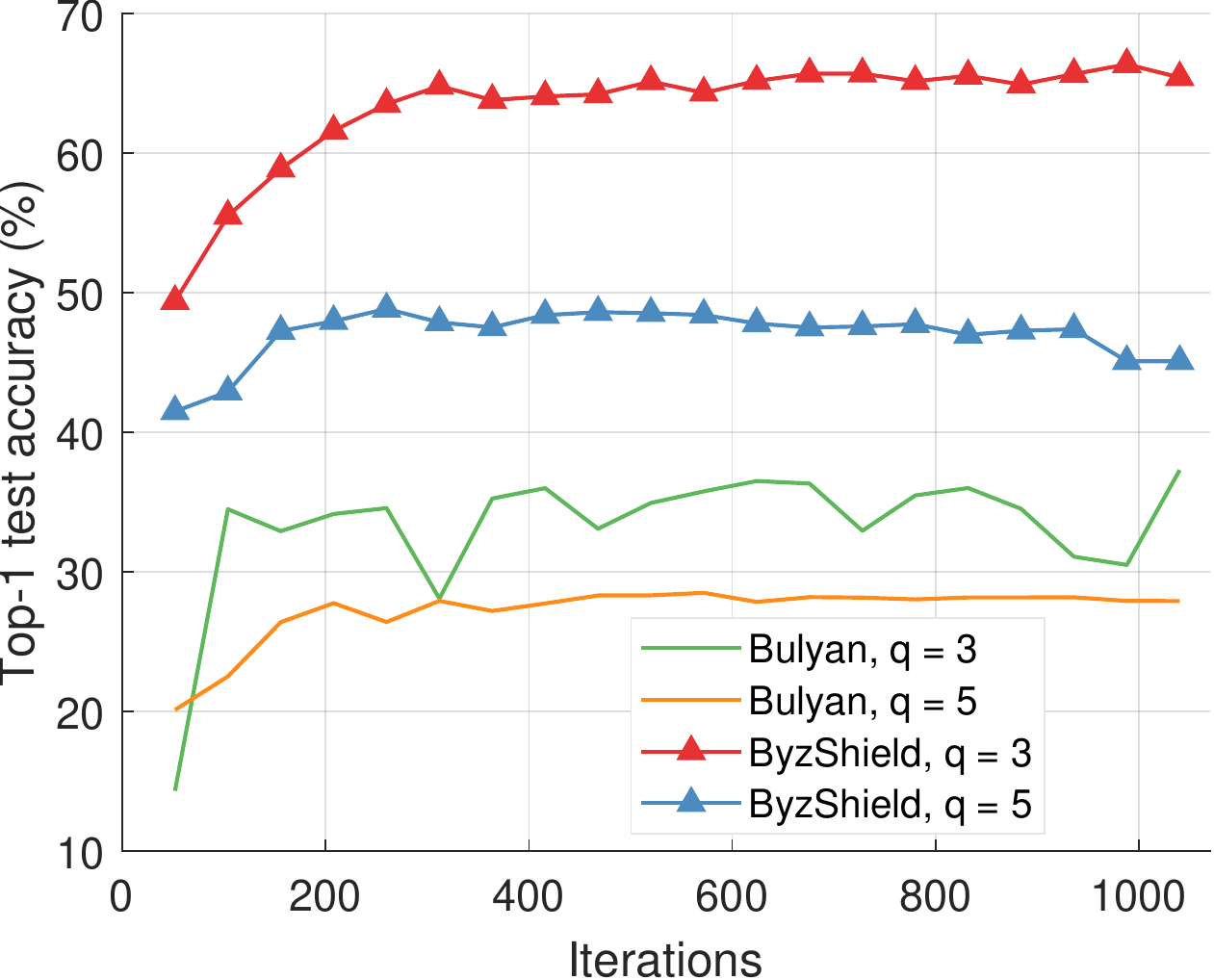}
	\caption{\emph{ALIE} attack and \emph{Bulyan}-based defenses (CIFAR-10).}
	\label{fig:top1_fig_63}
	\endminipage\hfill
	\minipage{0.32\textwidth}
	\includegraphics[width=\linewidth]{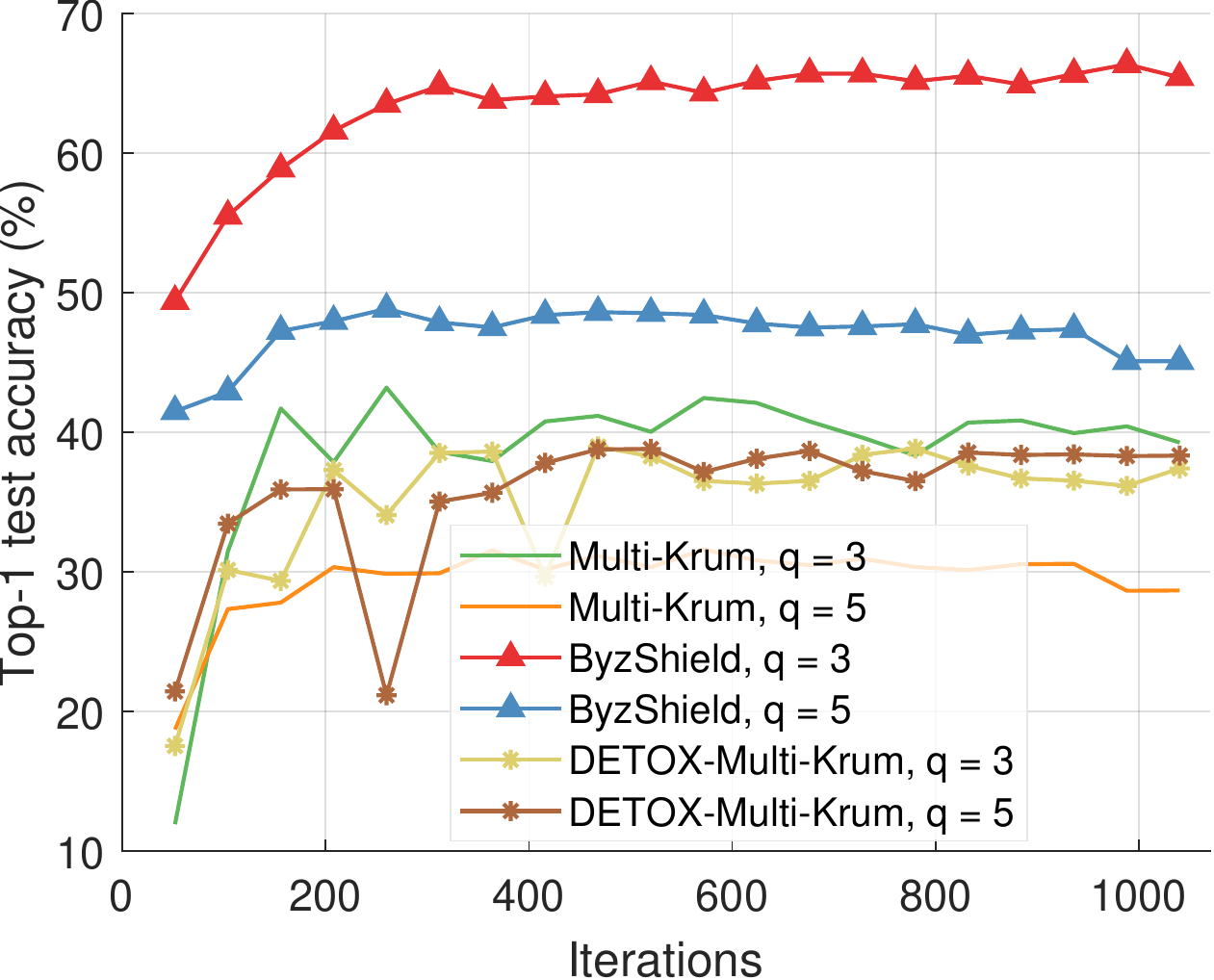}
	\caption{\emph{ALIE} attack and \emph{Multi-Krum}-based defenses (CIFAR-10).}
	\label{fig:top1_fig_64}
	\endminipage
	\vspace{-0.1in}
\end{figure*}

\begin{figure*}[!htb]
	\minipage{0.32\textwidth}
	\includegraphics[width=\linewidth]{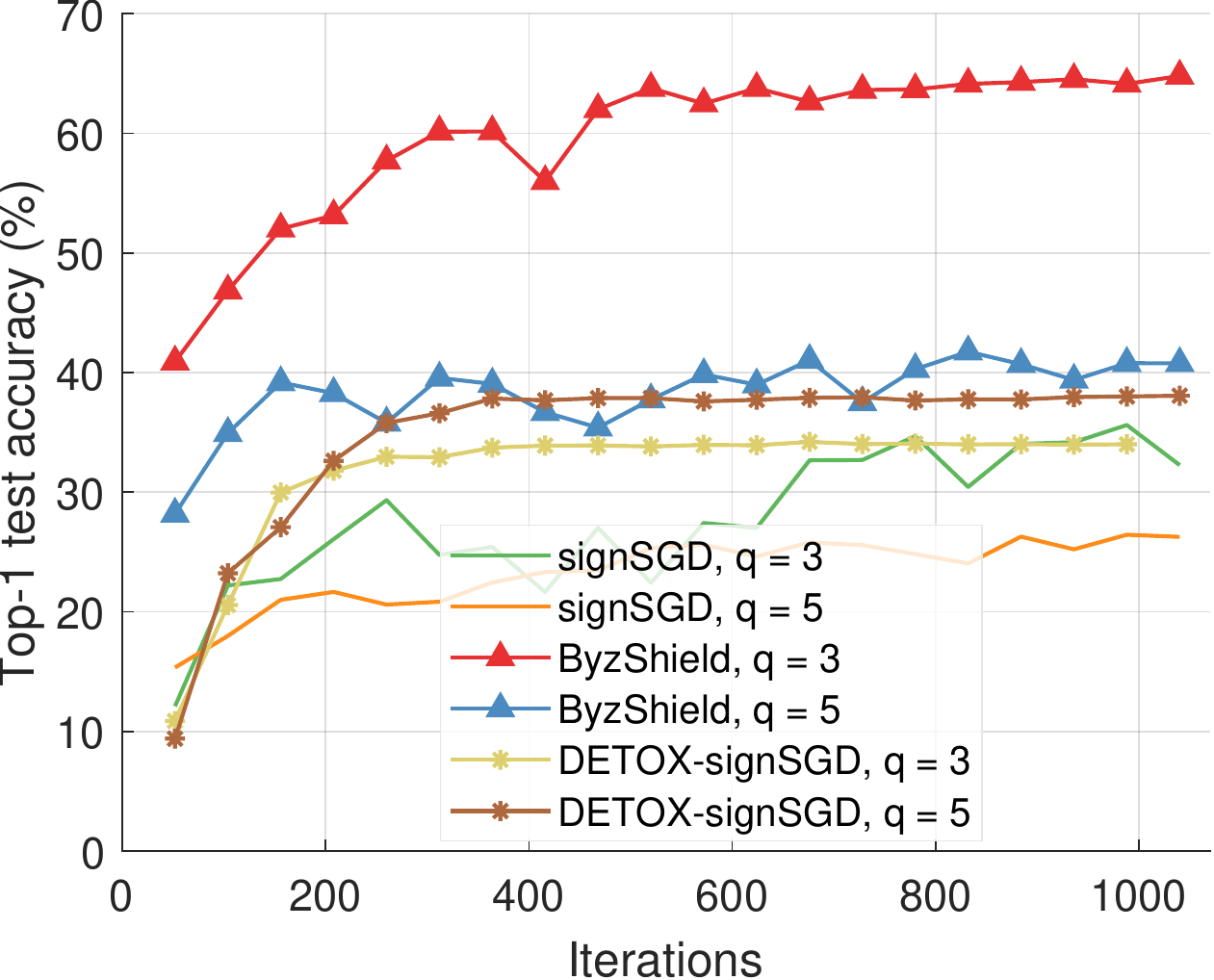}
	\caption{\emph{Constant} attack and \emph{signSGD}-based defenses (CIFAR-10).}
	\label{fig:top1_fig_65}
	\endminipage\hfill
	\minipage{0.32\textwidth}
	\includegraphics[width=\linewidth]{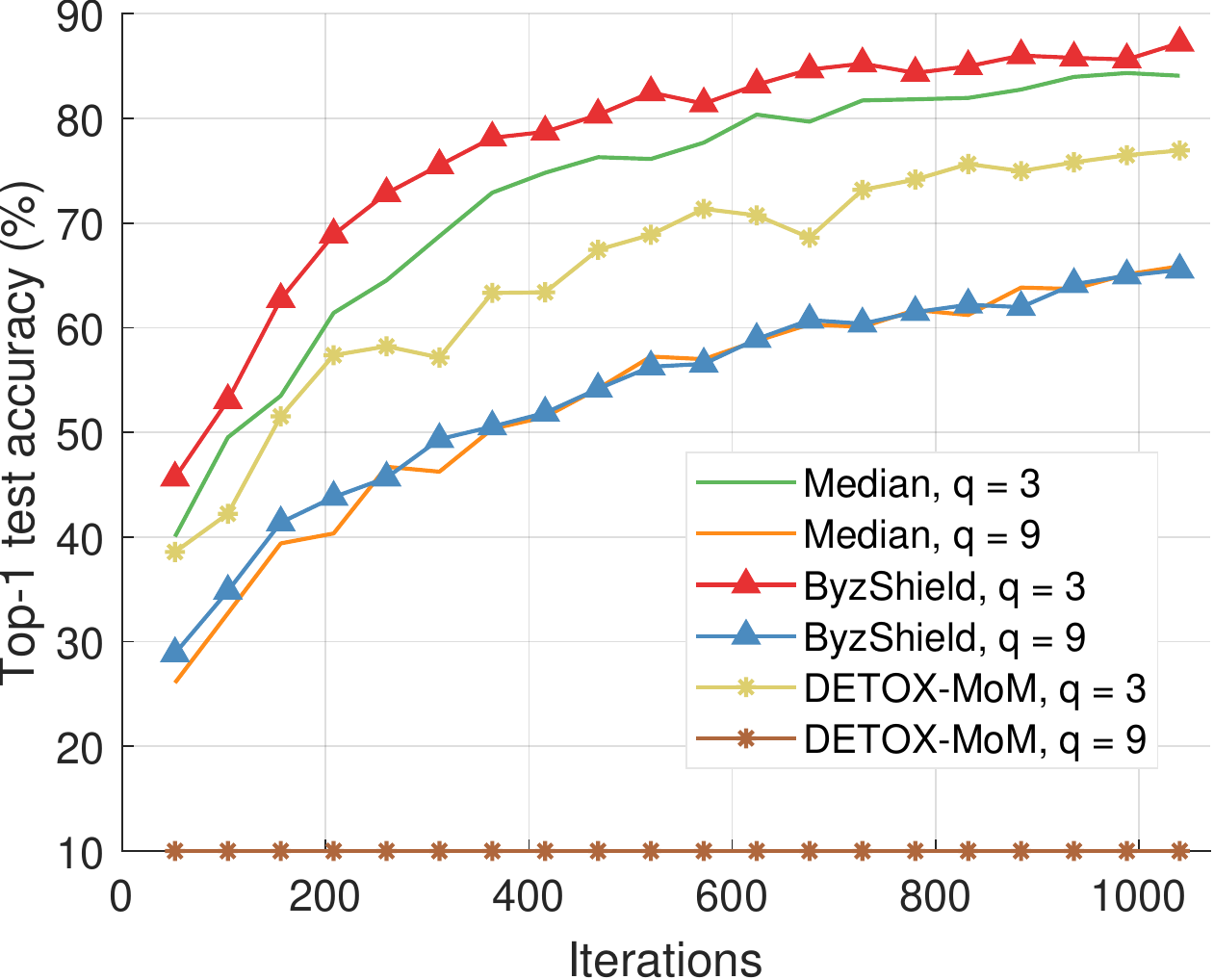}
	\caption{\emph{Reversed gradient} attack and median-based defenses (CIFAR-10).}
	\label{fig:top1_fig_66}
	\endminipage\hfill
	\minipage{0.32\textwidth}
	\includegraphics[width=\linewidth]{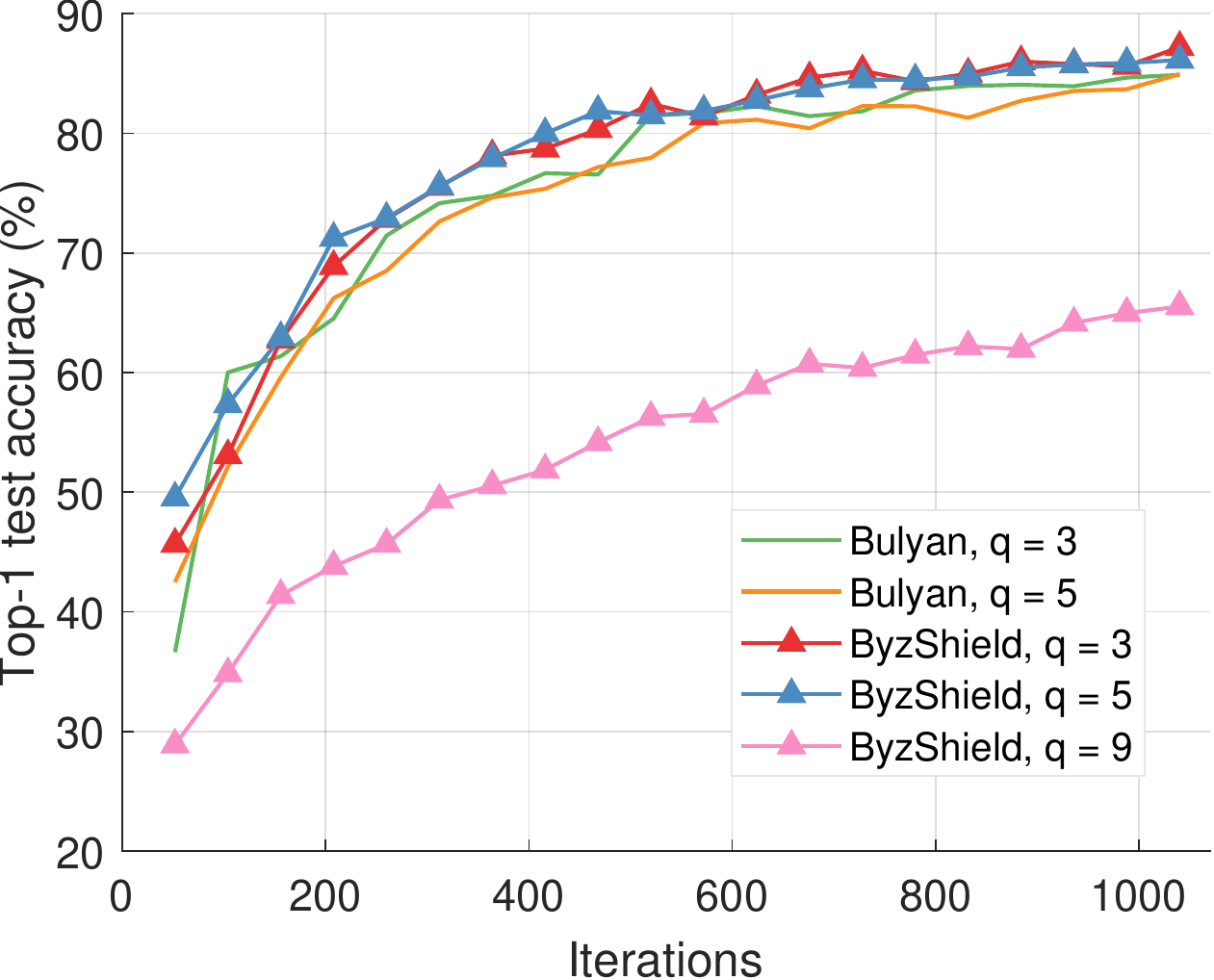}
	\caption{\emph{Reversed gradient} attack and \emph{Bulyan}-based defenses (CIFAR-10).}
	\label{fig:top1_fig_67}
	\endminipage
	\vspace{-0.1in}
\end{figure*}

\section{Large-scale Deep Learning Experiments}
\label{sec:experiments}
\subsection{Experiment Setup}
\label{sec:experiment_setup}
We have evaluated our method in classification tasks on large datasets on Amazon EC2 clusters. This project is written in PyTorch \cite{pytorch} and uses the MPICH library for communication among the devices. The implementation builds on DETOX's skeleton %\footnote{\href{https://github.com/hwang595/DETOX}{https://github.com/hwang595/DETOX}}
and has been provided.
%\footnote{\href{https://tinyurl.com/y2fcb6yp}{https://tinyurl.com/y2fcb6yp}}. 
The principal metric used in our evaluation is the top-1 test accuracy.

The classification tasks were performed using CIFAR-10 with ResNet-18 \cite{he_resnet}. Experimenting with hyperparameter tuning we have chosen the values for batch size, learning rate scheduling, and momentum, among others. We used clusters of $K=25$ and $K=15$ workers of type \texttt{c5.4xlarge} on Amazon EC2 for various values of $q$. The PS was an \texttt{i3.16xlarge} instance in both cases. More details are in the Appendix.

The cluster of $K=25$ workers utilizes the Ramanujan Case 2 construction presented in Section \ref{sec:ramanujan} with $r=l=5$ resulting in partitioning each batch into $f=rl=25$ files. For $K=15$, we used the MOLS scheme (\emph{cf.} Section \ref{sec:MOLS}) for $r=3$ and $l=5$ such that $f=l^2=25$. The corresponding simulated values of $c_{\mathrm{max}}^{(q)}$ and the fraction of distorted files, $\hat{\epsilon}$, under the attack on DETOX proposed in Section \ref{sec:simulations} are in Tables \ref{table:Ramanujan_distortion_test_l5r5} and \ref{table:MOLS_distortion_test_l5r3}, respectively.

The attack models we tried are the following.
\begin{itemize}
	\item \emph{ALIE} \cite{alie}: This attack involves communication among the Byzantines in which they jointly estimate the mean $\mu_i$ and standard deviation $\sigma_i$ of the batch's gradient for each dimension $i$. They subsequently send a function of those moments to the PS such that it will distort the median of the results.
	\item \emph{Constant}: Adversarial workers send a constant matrix with all elements equal to a fixed value; the matrix has the same dimensions as the true gradient.
	\item \emph{Reversed gradient}: Byzantines return $-c\*g$ for $c>0$, to the PS instead of the true gradient $\*g$.
\end{itemize}

The ALIE algorithm is, to the best of our knowledge, the most sophisticated attack in literature for centralized setups. Our experiments showcase this fact. Constant attack often alters the gradients towards wrong directions in order to disrupt the model's convergence and is also a powerful attack. Reversed gradient is the weakest among the three, as we will discuss in the Section \ref{sec:experiment_results}.

We point out that in all of our experiments, we chose a set of $q$ Byzantines (among all $\binom{K}{q}$ possibilities) such that $\hat{\epsilon}$ is maximized; this is equivalent to a worst-case omniscient adversary which can control any $q$ devices.

The defense mechanism ByzShield applied on the outputs of the majority votes is coordinate-wise median \cite{ramchandran_saddle_point} across the $f$ gradients computed as in Algorithm \ref{alg:main_algorithm} (lines \ref{alg_step:main_maj_vote_start}-\ref{alg_step:main_maj_vote_end}). We decided to pair our method with median since in most cases it worked well and we did not have to resort to more sophisticated baseline defenses.

\begin{figure}[t]
	\centering
	\includegraphics[scale=0.46]{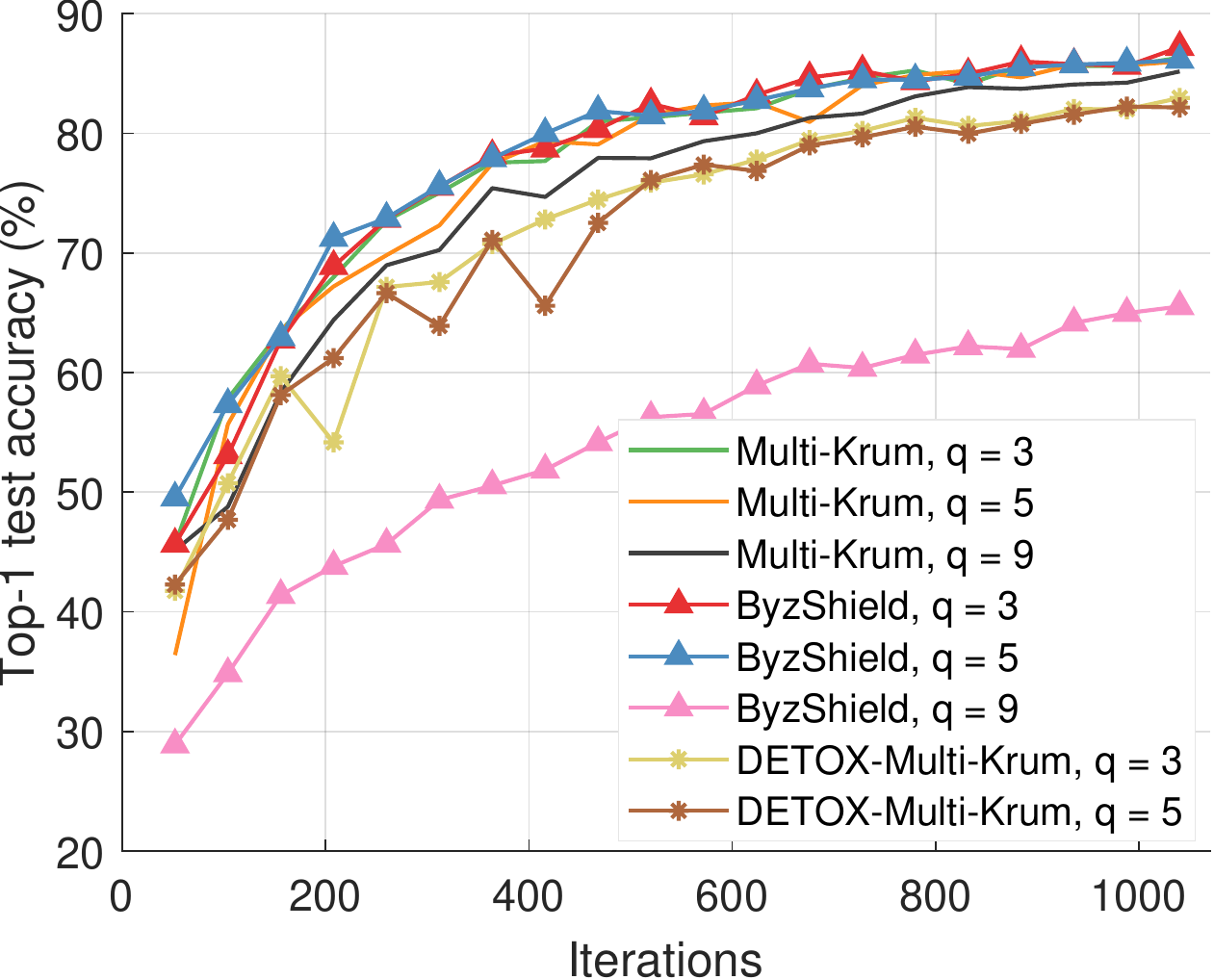}
	\caption{\emph{Reversed gradient} attack and \emph{Multi-Krum}-based defenses (CIFAR-10).}
	\label{fig:top1_fig_68}
	\vspace{-0.1in}
\end{figure}

The most popular baseline techniques we compare against are \emph{median-of-means} \cite{minsker2015}, coordinate-wise majority voting with \emph{signSGD} \cite{SIGNSGD}, \emph{Multi-Krum} \cite{aggregathor} and \emph{Bulyan} \cite{bulyan}. The latter two are robust to only a small fraction of erroneous results and their implementations rely on this assumption. Specifically, if $c_{\mathrm{max}}^{(q)}$ is the number of adversarial operands then Multi-Krum requires at least $2c_{\mathrm{max}}^{(q)}+3$ total number of operands while the same number for Bulyan is $4c_{\mathrm{max}}^{(q)}+3$. These constraints are very restrictive and the methods are inapplicable for larger values of $q$ for which our method is Byzantine-robust.
%All baseline techniques are applied directly on the $K$ values returned by the workers. 
DETOX \cite{detox} uses the aforementioned baseline methods for aggregating the ``winning" gradients after the majority voting.

\subsection{Results}
\label{sec:experiment_results}
In order to effectively interpret the results, the fraction of distorted gradients $\hat{\epsilon}$ needs to be taken into account; we will refer to the values in Appendix Table \ref{table:Ramanujan_distortion_test_l5r5} throughout this exposition. Due to the nature of gradient-descent methods, $\hat{\epsilon}$ may not be always proportional to how well a method works but it is a good starting point. We will restrict our interpretation of the results to the case of $K=25$. For $K=15$, we performed a smaller range of experiments demonstrating similar effects (\emph{cf.} Figures \ref{fig:top1_fig_69}, \ref{fig:top1_fig_70}, \ref{fig:top1_fig_71} in Appendix).

Let us focus on the results under the median-based defenses. In Figure \ref{fig:top1_fig_62}, we compare ByzShield, the baseline implementation of coordinate-wise median ($\hat{\epsilon}=0.12, 0.2$ for $q=3,5$, respectively) and DETOX with median-of-means ($\hat{\epsilon} = 0.2$) under the ALIE attack. ByzShield achieves at least a 20\% average accuracy overhead for both values of $q$ ($\hat{\epsilon}=0.04, 0.08$ for $q=3,5$, respectively). In Figure \ref{fig:top1_fig_66} (reversed gradient), we observe similar trends. However, for $q=9$ in DETOX, a fraction of $\hat{\epsilon}=0.6$ majorities is distorted and even though this attack is considered weak, the method breaks (constant $10$\% accuracy).

Similarly, ByzShield significantly outperforms Bulyan and Multi-Krum-based schemes on ALIE (Figures \ref{fig:top1_fig_63}, \ref{fig:top1_fig_64}); Bulyan cannot be paired with DETOX for $q\geq1$ for our setup since the requirement $f \geq 4c_{\mathrm{max}}^{(q)}+3$, as discussed in Section \ref{sec:experiment_setup}, cannot be satisfied. The maximum allowed $q$ for DETOX with Multi-Krum (Figures \ref{fig:top1_fig_64} and \ref{fig:top1_fig_68}) is $q=5$.

%\begin{table}[!t]
%	\centering
%	\caption{Total training time (in hours) for $q=3$ Byzantines.}
%	\label{table:time}
%	%\resizebox{0.8\columnwidth}{!}{
%	\setlength\tabcolsep{2pt}
%	\begin{tabular}{ |P{1.2cm}||P{1.4cm}|P{1.8cm}|P{1.5cm}| }
%		\hline
%		Figure & Baseline & ByzShield & DETOX \\
%		\hline
%		\ref{fig:top1_fig_62} & 3.14 & 10.81 & 4 \\
%		\ref{fig:top1_fig_63} & 4.04 & 10.81 & - \\
%		\ref{fig:top1_fig_64} & 3.36 & 10.81 & 3.88 \\
%		\ref{fig:top1_fig_65} & 0.87 & 7.35 & 19.78\\
%		\ref{fig:top1_fig_66} & 2.83 & 7.94 & 3.1\\
%		\ref{fig:top1_fig_67} & 3.27 & 7.94 & - \\
%		\ref{fig:top1_fig_68} & 3.06 & 7.94 & 2.84 \\
%		\hline
%	\end{tabular}
%	%}
%	%\vspace{-0.2in}
%\end{table}

In signSGD, only the sign information of the gradient is retained and sent to the PS by the workers. The PS will output the majority of the signs for each dimension. The method's authors \cite{SIGNSGD} suggest that gradient noise distributions are in principle unimodal and symmetric due to the \emph{Central Limit Theorem}. Following the advice of \cite{detox}, we pair this defense with the stronger constant attack as sign flips (\emph{e.g.,} reversed gradient) are unlikely to affect the gradient's distribution. ByzShield with median still enjoys an accuracy improvement of about 20\% for $q=3$ and a smaller one for $q=5$. The results are in Figure \ref{fig:top1_fig_65}. 

All schemes defend well under the reversed gradient attack for small values of $q$ (Figures \ref{fig:top1_fig_66}, \ref{fig:top1_fig_67}, \ref{fig:top1_fig_68}). Nevertheless, in Figure \ref{fig:top1_fig_67}, by testing ByzShield on large number of Byzantine workers ($q=9$) for which a fraction of $\hat{\epsilon}=0.36$ files are distorted, we realize that it still converges to a relatively high accuracy; Bulyan cannot be applied in this case. Multi-Krum has an advantage over our method for this attack for $q=9$ (Figure \ref{fig:top1_fig_68}). DETOX cannot be paired with Multi-Krum in this case as it needs at least $2c_{\mathrm{max}}^{(q)}+3=7$ groups.

Note that in our adversarial scenario, DETOX may perform worse than its baseline counterpart depending on the fraction of corrupted gradients. For example, in Figure \ref{fig:top1_fig_66}, for $q=3$, $\hat{\epsilon}^{DETOX}=0.2$ while $\hat{\epsilon}^{Median}=3/25=0.12$ and this reflects on better accuracy values for the baseline median.

There are two takeaways from these results. First, the distortion fraction $\hat{\epsilon}$ is a good predictor of an aggregation scheme's convergence. Second, ByzShield supersedes prior schemes by an average of $20$\% in terms of top-$1$ accuracy for large values of $q$ under various powerful attacks.
%\begin{itemize}
%	\item The distortion fraction $\hat{\epsilon}$ is a good predictor of an aggregation scheme's convergence.
%	\item ByzShield supersedes prior schemes by an average of 20\% in terms of top-1 accuracy for large values of $q$ under various powerful attacks.
%\end{itemize}

\textbf{Training Time}: 
%We next discuss computation and communication time issues.
In ByzShield, within an iteration, a worker performs $l$ forward/backward propagations %\aditya{potential confusion between this ell and other ell} 
(one for each file) and transmits $l$ gradients back to the PS. The rest of the approaches have each worker return a single gradient. Hence, ByzShield spends more time on communication. In both DETOX and ByzShield a worker processes $r$ times more samples compared to a baseline technique. 
%Nevertheless, only one gradient is computed and sent to the PS according to DETOX; this is because each worker in a group processes exactly one file. 
Thus, we expect redundancy-based methods to consume more time on computation at the worker level than their baseline counterparts. Aggregation time varies a lot by the method and the number of files to be processed by the PS for the model update. 
As an example, baseline median, ByzShield and DETOX median-of-means took 3.14, 10.81 and 4 hours, respectively, for the full training under the ALIE attack with $q=3$ (Figure \ref{fig:top1_fig_62}). The corresponding per-iteration time (average across iterations) is in Figure \ref{fig:ALIE_median_per_iteration} of the Appendix.
%We present the total training time for $q=3$ for our experiments in Figures \ref{fig:top1_fig_62}-\ref{fig:top1_fig_68} in Table \ref{table:time}. 

%We emphasize though that the central finding of our work is that careful assignment of tasks to workers can significantly improve the overall robustness of the training and our focus was not on optimizing the implementation. There is ample scope for improving the implementation to reduce the overall time consumption. This can be done, e.g., by using GPUs or workers with higher network bandwidth.
%A detailed time analysis is not the focus of this work. We emphasize that we are not experts in software optimizations and we did not focus too much on this aspect for our Python implementation. Such optimizations (e.g., adding support for GPUs) as well as using a less time-consuming robust aggregation after the majority voting can decrease training time.
%The conclusions that we can draw for Multi-Krum from Figures \ref{fig:top1_fig_41} and \ref{fig:top1_fig_43} are very similar.

\section{Conclusions and Future Work}
We chose to utilize median for our method as it was performing well in all of our experiments.
% and consistently achieve convergence to higher accuracy values under the most sophisticated Byzantine attacks, compared to other popular defenses. 
ByzShield can also be used with non-trivial aggregation schemes such as Bulyan and Multi-Krum and potentially yield even better results. Our scheme is significantly more robust than the most sophisticated defenses in literature at the expense of increased computation and communication time. We believe that implementation-related improvements such as adding support for GPUs can alleviate the overhead. Algorithmic improvements to make it more communication-efficient are also worth exploring. Finally, we considered an omniscient adversary and the worst-case choice of Byzantines. We think that this is a reasonable method if one wants to make fair and consistent comparisons across different schemes. But, weaker random attacks may also be an interesting direction.

\section*{Acknowledgments}
This work was supported in part by the National Science Foundation (NSF) under grant CCF-1718470 and grant CCF-1910840.

\clearpage

\bibliographystyle{mlsys2021}
%\nocite{*}
\bibliography{citations}
%{\small\bibliography{./bib/citations}}

%%%%%%%%%%%%%%%%%%%%%%%%%%%%%%%%%%%%%%%%%%%%%%%%%%%%%%%%%%%%%%%%%%%%%%%%%%%%%%%
%%%%%%%%%%%%%%%%%%%%%%%%%%%%%%%%%%%%%%%%%%%%%%%%%%%%%%%%%%%%%%%%%%%%%%%%%%%%%%%
% SUPPLEMENTAL CONTENT AS APPENDIX AFTER REFERENCES
%%%%%%%%%%%%%%%%%%%%%%%%%%%%%%%%%%%%%%%%%%%%%%%%%%%%%%%%%%%%%%%%%%%%%%%%%%%%%%%
%%%%%%%%%%%%%%%%%%%%%%%%%%%%%%%%%%%%%%%%%%%%%%%%%%%%%%%%%%%%%%%%%%%%%%%%%%%%%%%
\clearpage
\appendix
\section{Appendix}

\subsection{Asymptotic Complexity}
\label{supplementary:asymptotic}
%\aditya{I think this asymptotic complexity discussion can be moved to the Appendix since it doesn't appear relevant even later}
%\textbf{Asymptotic complexity}:
If the gradient computation has linear complexity and since each worker is assigned to $l$ files of $b/f$ samples each, the gradient computation cost at the worker level is $\mathcal{O}((lb/f)d)$ ($K$ such computations in parallel). In our schemes, however, $b$ is a constant multiple of $f$ and $l \leq r$; hence, the complexity becomes $\mathcal{O}(rd)$ which is identical to other redundancy-based schemes in \cite{detox, draco}. The complexity of aggregation varies significantly depending on the operator. For example, majority voting can be done in time which scales linearly with the number of votes using \emph{MJRTY} proposed in \cite{boyer_majority}. In our case, this is $\mathcal{O}(Kd)$ as the PS needs to use the $d$-dimensional input from all $K$ machines. Krum \cite{blanchard_krum}, Multi-Krum \cite{aggregathor} and Bulyan \cite{bulyan} are applied to all $K$ workers by default and require $\mathcal{O}(K^2(d+\mathrm{log}K))$.

%\subsection{Proof of Equation \eqref{eq:tanner_vol_expansion_workers_files}}
%\begin{proof}
%	By Lemma \ref{lemma:tanner_vol_expansion}
%	\begin{eqnarray*}
%		&&\frac{\mathrm{vol}(\calN(S))}{\mathrm{vol}(S)} \geq \frac{1}{\mu_1 + (1-\mu_1)\frac{\mathrm{vol}(S)}{|\calE|}}\\ \Rightarrow&& \frac{\mathrm{vol}(\calN(S))}{ql} \geq \frac{1}{\mu_1 + (1-\mu_1)\frac{ql}{Kl}}\\
%		\Rightarrow&& \mathrm{vol}(\calN(S)) \geq \frac{ql}{\mu_1 + (1-\mu_1)\frac{q}{K}}\\
%		\Rightarrow&& |\calN(S)|r \geq \frac{ql}{\mu_1 + (1-\mu_1)\frac{q}{K}}\\
%		\Rightarrow&& |\calN(S)| \geq \frac{ql/r}{\mu_1 + (1-\mu_1)\frac{q}{K}}.
%	\end{eqnarray*}
%\end{proof}

%\subsection{Proof of Claim \ref{claim:tanner_average_degree}}
%\begin{proof}
%	Based on the volume of $S$ the average degree is less than or equal to
%	\begin{equation*}
%		\frac{|S|l}{\beta} = \left(\mu_1 + (1-\mu_1)\frac{q}{K}\right)r
%	\end{equation*}
%	based on Observation \ref{claim:tanner_vol_expansion_workers_files}.
%\end{proof}

\subsection{Proof of Claim \ref{claim:max_distortion_matching_attack}}
\begin{proof}
As all nodes in $S$ have degree $l$, the number of outgoing edges from $S$ is $l|S|$. For a file to be distorted it needs at least $r'$ of its copies to be processed by Byzantine nodes. Furthermore, we know that collectively the nodes process at least $\beta$ files. Using these observations, we have
	\begin{eqnarray}
		&&l|S| \geq c_{\mathrm{max}}^{(q)}r' + (|\calN(S)| - c_{\mathrm{max}}^{(q)})\nonumber\\
		\Rightarrow&&l|S| \geq c_{\mathrm{max}}^{(q)}r' + (\beta - c_{\mathrm{max}}^{(q)})\label{eq:max_distortion_matching_attack_boundary}\\
		\Rightarrow&& l|S| \geq c_{\mathrm{max}}^{(q)}(r'-1)+\beta\nonumber\\
		\Rightarrow&& l|S| - \beta \geq c_{\mathrm{max}}^{(q)}(r'-1)\nonumber\\
		\Rightarrow&& c_{\mathrm{max}}^{(q)} \leq \frac{l|S| - \beta}{r'-1} = \frac{ql-\beta}{(r-1)/2} \defeq \gamma \nonumber
	\end{eqnarray}
	where in Eq. \eqref{eq:max_distortion_matching_attack_boundary} we have substituted the result from Eq. \eqref{eq:tanner_vol_expansion_workers_files}.
\end{proof}

\subsection{Proof of Lemma \ref{lemma:spectrum_of_all_schemes}}
\label{supplementary:spectrum_of_all_schemes}
\begin{proof}
	From \cite{ramanujan_bigraphs_Shantanu} [Theorem 6], the singular values of the zero-one (unnormalized) biadjacency matrix $H$ can be characterized in three distinct cases; we utilize the first two of them. They are presented below with the singular values reported in decreasing order of magnitude.
	
	\begin{itemize}
		\item \textbf{Ramanujan Case 1}: If $2 \leq m < s$, then $H=B^T$ has singular values and corresponding multiplicities
		$$\{(\sqrt{sm},1), (\sqrt{s},m(s-1)), (0,m-1)\}.$$
		\item \textbf{Ramanujan Case 2}: If $m \geq s$ and $s|m$, then $H=B$ has singular values and corresponding multiplicities 
		$$\{(\sqrt{sm},1), (\sqrt{m},(s-1)s), (0,s-1)\}.$$
		%	\item \textbf{Case 3}: If $m \geq s$ and $s \nmid m$, let $k=m \text{ mod } s$, then $H=B$ has singular values and corresponding multiplicities 
		%	$$\{(\sqrt{sm},1), (\sqrt{m+s-k},(s-1)k), (\sqrt{m-k},(s-1)(s-k)), (0,s-1)\}$$
	\end{itemize}
	
	In order to perform a direct comparison between the MOLS and Ramanujan construction we want to examine the eigenvalues of the normalized biadjacency matrix $A=\frac{1}{\sqrt{d_Ld_R}}H$ (from Section \ref{sec:tanner_expansion_resilience}) multiplied by its transpose, i.e., we are interested in the spectrum of $(AA^T)_{Ram. 1}$ and $(AA^T)_{Ram. 2}$. In all cases, observe that $d_Ld_R=sm$ and the spectrum of $AA^T=\frac{1}{sm}HH^T$ in Lemma \ref{lemma:spectrum_of_all_schemes} of the main text is easily deduced; these eigenvalues will be the squares of the singular values of $H$ multiplied by $\frac{1}{sm}$.
	
	We continue by computing the spectrum of $(AA^T)_{MOLS}$. For a given Latin square $L$ we will use the term \emph{parallel class} to denote the collection, i.e., the multiset consisting of the sets $\mathcal{N}(U_{i_1}), \mathcal{N}(U_{i_2}), \dots, \mathcal{N}(U_{i_l})$, where $U_{i_1}, U_{i_2}, \dots, U_{i_l}$ are the workers populated based on $L$. The classes for Latin squares $L_1,\dots,L_r$ will be denoted with $\mathcal{P}_1,\dots,\mathcal{P}_r$, respectively. The underlying principle is that a parallel class includes exactly one replica of each file. Let 
	\begin{equation*}
		X = l\left(\frac{1}{\sqrt{lr}}\right)^2I_l = \frac{1}{r}I_l =
		\left(
		\begin{array}{ccccc}
			1/r& & & & \bigzero \\
			& 1/r & & & \\
			& & \ddots & & \\
			\bigzero & & & & \\
			& & & & 1/r \\
		\end{array}\right)
	\end{equation*}
	and
	\begin{equation*}
		Y = 1\left(\frac{1}{\sqrt{lr}}\right)^2J_l = \frac{1}{lr}J_l
	\end{equation*}
	where $J_n$ is the all-ones $n \times n$ matrix.
	
	Based on the placement scheme introduced in Section \ref{sec:allocation} the matrix $(AA^T)_{MOLS}$ takes the following $K \times K$ block form.
	\begin{equation*}
		(AA^T)_{MOLS}=
		\begin{bmatrix} 
			X & Y & Y & \cdots & \cdots & Y \\ 
			Y & X & Y & Y & \cdots & Y \\ 
			Y & Y & X  & Y & \ddots & \vdots \\ 
			\vdots & \ddots & \ddots & \ddots &  \ddots & Y \\ 
			Y & \cdots & \cdots & Y & X & Y \\ 
			Y & \cdots & \cdots & \cdots & Y & X 
		\end{bmatrix}
	\end{equation*}
	
	Hence, the blocks of $(AA^T)_{MOLS}$ take one of the forms $X,Y$. This is shown by observing that blocks $X$ concern multiplication of rows of $A$ with columns of $A^T$ which correspond to workers within a parallel class (if the index of the row of $A$ is the same as the index of the column of $A^T$ then $l$ entries $(\frac{1}{\sqrt{lr}})^2$ are added, otherwise there is no common file and the product is zero). Similarly, blocks that have the form of $Y$ concern pairs of workers from different parallel classes which share exactly one file.
	
	We observe that
	\begin{equation}
		\label{eq:M_MOLS_def}
		(AA^T)_{MOLS} = \frac{1}{lr}
		\underbrace{
			\begin{pmatrix}
				0&1&\cdots & 1\\
				1&0&\ddots&\vdots\\
				\vdots&\ddots&\ddots&1\\
				1&\cdots&1&0
			\end{pmatrix}
		}_{=\ C_{r \times r}} \otimes J_l + \frac 1r I_{lr}.
	\end{equation}
	
	The characteristic polynomial of $J_l$ is $(\lambda-l)\lambda^{l-1}$ and its spectrum (with corresponding algebraic multiplicities) is $\sigma(J_l) = \{(l,1), (0,l-1)\}$ \cite{horn_matrix_analysis} which coincides with the spectrum of the adjacency matrix of a complete graph with self-edges.
	
	\begin{note}
		\label{note:rank1_update}
		If $\lambda$ is an eigenvalue of $A$ with eigenvector $\mathbf{x}$, then $\lambda+c$ is an eigenvalue of $A+cI$ with the same eigenvector $\mathbf{x}$.
	\end{note}
	
	For $C$, note that $\sigma(C) = \{(r-1,1),(-1,r-1)\}$ since it is a rank-1 update of $J_r$ (\emph{cf.} Note \ref{note:rank1_update}).
	
	Let $D \coloneqq C \otimes J_l$ and spectra $\sigma(C)=\{\lambda_1,\dots,\lambda_r\}$ and $\sigma(AA^T)=\{\mu_1,\dots,\mu_l\}$ then
	\begin{eqnarray*}
		\sigma(D)&=&\{\lambda_i\mu_j, i=1,\dots,r,j=1,\dots,l\}\\
		&=&\{(l(r-1),1),(-l,r-1),(0,(l-1)r)\}.
	\end{eqnarray*}
	
	\begin{note}
		\label{note:scalar_update}
		If $\lambda$ is an eigenvalue of $A$ with eigenvector $\mathbf{x}$, then $\alpha\lambda$ is an eigenvalue of $\alpha A$ with the same eigenvector $\mathbf{x}$.
	\end{note}
	
	Using Note \ref{note:scalar_update}, the spectrum of the term $\frac{1}{lr}D$ in Eq. \eqref{eq:M_MOLS_def} becomes
	\begin{eqnarray*}
		&\left\{\left(\frac{1}{lr}l(r-1), 1\right), \left(\frac{1}{lr}(-l), r-1\right), \left(\frac{1}{lr}0, (l-1)r\right)\right\}&\\
		&= \left\{\left(\frac{1(r-1)}{lr}, 1\right), (-1/r, r-1), (0, (l-1)r)\right\}&
	\end{eqnarray*}
	
	which by Note \ref{note:scalar_update} yields
	\begin{eqnarray*}
		\sigma((AA^T)_{MOLS}) &=& \Big\{\Big(\frac{1(r-1)}{lr}+1/r, 1\Big), \cdots\\
		&&\hspace{8pt} (-1/r+1/r, r-1), \cdots\\
		&&\hspace{8pt} (0+1/r, (l-1)r)\Big\}\\
		&=& \{(1, 1), (0, r-1), (1/r, (l-1)r)\}.
	\end{eqnarray*}
\end{proof}

\subsection{Proof of Claim \ref{claim:r_less_q_distortions}}
\begin{proof}
	We begin by proving the result for the MOLS-based scheme. We note that there is a bijection between the files and the grid's cells $(i,j)$ (\emph{cf.} Section \ref{sec:MOLS}); we will refer to the files using their cell's coordinates and vice versa. A worker $U_k$ which belongs to $L_\alpha$ is such that its files are completely characterized by the solutions to an equation of the form $\alpha i + j = z$. 
	
	These facts immediately imply that workers belong to a LS $L_\alpha$ do not have any files in common. As the equations corresponding to $L_\alpha$ and $L_\beta$ for $\alpha \neq \beta$ are linearly independent, we also have that any two workers from different Latin squares share exactly one file.
	
	The respective proofs for the two cases of the claim are given below.
	\begin{itemize}
		\item \textbf{Case 1}: $r=3$. 
		
		For $q=2$ and based on the above observations, it is evident that attacking $q=2$ workers cannot distort more than one file, their common file. This yields $c_{\mathrm{max}}^{(q)}=1$ and $\hat{\epsilon} = 1/f$. 
		
		For $q=3$, we want to show that we can distort up to $c_{\mathrm{max}}^{(q)}=3$ files. Suppose that the three Latin squares in which the workers lie are denoted $L_{\alpha_i}, i = 1, 2, 3$ where the $\alpha_i$'s are distinct. We can choose any $z_1$ and $z_2$ such that the workers $U_1$ and $U_2$ specified by $\alpha_1 i + j = z_1$ and $\alpha_2 i + j = z_2$ respectively. Suppose that the common file in these workers is denoted by $(i_1, j_1)$. For $U_3$ specified by $\alpha_3 i + j = z_3$ we only need to ensure that we pick $z_3 \neq \alpha_3 i_1 + j_1$. This can always be done since there are $l-1$ other choices for $z_3$.
		
		With these we can see that any two workers within the set $\{U_1, U_2, U_3\}$ share a file. As $r=3$, this implies that $c_{\mathrm{max}}^{(q)} = 3$ for $q=3$ and $\hat{\epsilon} = 3/f$.
		\item \textbf{Case 2}: $r>3$. 
		
		The argument we make for this case follows along the same lines as that for $q=3$ above. Note that to distort a given file we have to pick  $r'$ workers $U_{k_1},\dots,U_{k_{r'}}$ from distinct LS $L_{\alpha_1},\dots,L_{\alpha_{r'}}$ which correspond to a system of equations of the form 
		\begin{equation}
			\label{eq:first_r_prime_system}
			\begin{array}{lll}
				\alpha_1 i+j &=& z_{k_1}\\
				\alpha_2 i+j &=& z_{k_2}\\
				&\vdots&\\
				\alpha_{r'} i+j &=& z_{k_{r'}}.
			\end{array} 
		\end{equation}
		This system of equations has to have a common solution denoted by $(i_1, j_1)$. We can pick at most $r'-1$ more workers $U_{k_{r'+1}},\dots,U_{k_{r}}$ since $q \leq r$. If these workers have to distort a different file, then they need to certainly have a common solution denoted $(i_2, j_2)$ along with one worker from $\{U_{k_1},\dots,U_{k_{r'}}\}$. Note that we cannot pick two workers from  $\{U_{k_1},\dots,U_{k_{r'}}\}$ since any two such workers already share $(i_1, j_1)$.
		
		Finally, we note that among these workers we cannot find another subset of $r'$ workers that share a third file $(i_3, j_3)$ since any such subset will necessarily include at least two files from either $\{U_{k_1},\dots,U_{k_{r'}}\}$ or $\{U_{k_{r'+1}},\dots,U_{k_{r}}\}$. This concludes the proof.
		
	\end{itemize}

	We continue by proving the claim for Ramanujan Case 1. The structure of the biadjacency matrix $B$ in Section \ref{sec:ramanujan} will reveal useful properties used in this proof. Since Ramanujan Case 2 uses $B^T$ instead of $B$, that proof is very similar and will be omitted.
	
	Let us start by substituting the parameters for Case 1 which are $m=r$ and $s=l$ so that $B$ becomes an $l \times r$ block matrix with dimensions $l^2\times rl$ (tall matrix)
	\begin{equation}
		\label{eq:ramanujan_C1_B}
		B=
		\begin{bmatrix}
			I_{l} & I_{l} & I_{l} & \cdots & I_{l}\\
			I_{l} & P & P^2 & \cdots & P^{r-1}\\
			I_{l} & P^2 & P^4 & \cdots & P^{2(r-1)}\\
			I_{l} & P^3 & P^6 & \cdots & P^{3(r-1)}\\
			\vdots & \vdots & \vdots & \vdots & \vdots\\
			I_{l} & P^{l-1} & P^{2(l-1)} & \cdots & P^{(r-1)(l-1)}\\
		\end{bmatrix}.
	\end{equation}

	As in the case of MOLS, suppose that the workers are partitioned into $r$ parallel classes $\mathcal{P}_1,\dots,\mathcal{P}_r$ depending on which block-column of $B$ was used to populate them. In this case, the $i^{th}$ block-column of $B$ is mapped to the $i^{th}$ parallel class of workers, $\mathcal{P}_i$.
	
	Also, define a bijection $\phi$ between the indices ${0,\dots,l^2-1}$ of the files (rows of $B$) and tuples of the form $(x,y)$ where $x,y\in\{0,\dots,l-1\}$ such that the mapping for a file index $i$ is
	\begin{equation*}
		\phi:\ i \rightarrow (x,y) = (\lfloor i/l\rfloor,\ i\ (\mathrm{mod}\ l)).
	\end{equation*}
	In the remaining part, we will be referring to the files as tuples $(x,y)$. %With a similar mapping, we map each worker $U_k$ to an integer $\chi(U_k) \in \{0,\dots,l-1\}$ such that
%	\begin{equation*}
%		\chi:\ U_k \rightarrow \lfloor k/r\rfloor.
%	\end{equation*}

	Our next objective is to describe each worker $U_k$ with a linear equation which encodes the relationship between $x$ and $y$ of the files $(x,y)$ assigned to $U_k$. 
	
	Due to its special structure, let us focus on the first block column which corresponds to $\mathcal{P}_1$. Note that the files stored by each of the first $l$ workers $U_0,\dots,U_{l-1}$ respectively satisfy the following equations
	\begin{equation}
		\label{eq:ramanujan_C1_first_class}
		\begin{array}{lll}
			y &=& 0\ (\mathrm{mod}\ l)\\
			y &=& 1\ (\mathrm{mod}\ l)\\
			&\vdots&\\
			y &=& l-1\ (\mathrm{mod}\ l).
		\end{array}
	\end{equation}
	We want to derive the equations for the remaining parallel classes $\mathcal{P}_2,\dots,\mathcal{P}_r$. For a class $\mathcal{P}_i\in\{\mathcal{P}_2,\dots,\mathcal{P}_r\}$ and letting $U_{k_1},\dots,U_{k_{l-1}}$ be the workers of $\mathcal{P}_i$, their files respectively satisfy
	\begin{equation}
		\label{eq:ramanujan_C1_remaining_classes}
		\begin{array}{lll}
			(i-1)x-y &=& 0\ (\mathrm{mod}\ l)\\
			(i-1)x-y &=& 1\ (\mathrm{mod}\ l)\\
			&\vdots&\\
			(i-1)x-y &=& l-1\ (\mathrm{mod}\ l).
		\end{array}
	\end{equation}
	for all values of $i\in \{2,\dots,r\}$. Simple observations of the structure of powers of the permutation matrix $P$ of Step 2 in Section \ref{sec:ramanujan} and that of matrix $B$ in Eq. \eqref{eq:ramanujan_C1_B} should convince the reader of the above fact.
	
	Let us proceed to the optimal distortion analysis. The ensuing discussion is almost identical to that for MOLS presented earlier in the current section (refer to that for more details).
	\begin{itemize}
		\item \textbf{Case 1}: $r=3$.
		
		For $q=2$, it is straightforward to see that any pair of linearly independent equations (from two adversarial workers of different classes) yields their unique common file $(x,y)$ which is distorted; hence $c_{\mathrm{max}}^{(q)}=1$ and $\hat{\epsilon} = 1/f$.
		
		For $q=3$, we can either pick one equation (worker) from Eq. \eqref{eq:ramanujan_C1_first_class} and the remaining two from Eq. \eqref{eq:ramanujan_C1_remaining_classes} or pick all three from Eq. \eqref{eq:ramanujan_C1_remaining_classes}. In any case, an appropriate choice of the three equations can guarantee that any subset of two of them are linearly independent so that $c_{\mathrm{max}}^{(q)}=3$ gradients are distorted and $\hat{\epsilon} = 3/f$.
		
		\item \textbf{Case 2}: $r>3$.
		
		Pick the first $r'$ adversaries $U_{k_1},\dots,U_{k_{r'}}$ from distinct classes (the choice of classes is irrelevant). Jointly solving any two of them  fully describes a gradient $(x,y)$ and substituting that solution to the remaining equations will guide us to the choice of the remaining Byzantines needed to distort the file $(x,y)$.
				
		Then, for the other $r'-1$ Byzantines we make sure that at least one of them, say $U_k$, does not belong to any of the classes which include $U_{k_1},\dots,U_{k_{r'}}$ and that $U_k$ does not store $(x,y)$; this implies that by solving the system of the equations of $U_k$ and any other Byzantine chosen so far we compute a file $(x',y')$. To skew $(x',y')$ we need $r'-2$ new adversaries $U_{k_{r'+1}},\dots,U_{k_{r}}$ which store it. It is clear that $(x',y')$ has to be a solution to their equations (none of these adversaries can be in $\{U_{k_1},\dots,U_{k_{r'}}\}$). Then, $c_{\mathrm{max}}^{(q)}=2$ and $\hat{\epsilon} = 2/f$.
	\end{itemize}
\end{proof}

\subsection{Extra Simulation Results on $c_{\mathrm{max}}^{(q)}$ and $\hat{\epsilon}$}
\label{supplementary:extra_e_hat_simulations}
More simulations similar to those of Section \ref{sec:simulations} of the main paper are shown in Tables \ref{table:Ramanujan_distortion_test_l5r5}, \ref{table:MOLS_distortion_test_l7r5} and \ref{table:MOLS_distortion_test_l7r3}. Note that we evaluate only up to $q=13$ in Table \ref{table:MOLS_distortion_test_l7r5} since the problem quickly becomes computationally intractable and takes too long to finish as the number of combinations scales exponentially.

\subsection{Deep Learning Experiments Hyperparameters}
The images have been normalized using standard values of mean and standard deviation for the dataset.  We chose $b=750$ for both cases ($K=25$ and $K=15$). The value used for momentum was set to $0.9$ and we trained for $13$ epochs in all experiments.  In Table \ref{table:tuning}, a learning rate scheduling is denoted with $(x,y,z)$; this notation signifies the fact that we start with a rate equal to $x$ and every $z$ iterations we multiply the rate by $y$. We will also index the schemes in order of appearance in the corresponding figure's legend. Experiments of ByzShield which appear in multiple figures are not repeated here (we ran those training processes once). In order to pick the optimal hyperparameters for each scheme, we performed an extensive grid search involving different combinations of $(x,y,z)$. For each method, we ran 200 iterations for each such combination and chose the one which was giving the lowest value of average cross-entropy loss (principal criterion) and the highest value of top-1 accuracy (secondary criterion).

\subsection{Software}
Our implementation of the ByzShield algorithm used for the experiments (Section \ref{sec:experiments}) as well as the simulations of the worst-case attacks (Section \ref{sec:simulations}) is available at \href{https://github.com/kkonstantinidis/ByzShield}{https://github.com/kkonstantinidis/ByzShield}. The implementation of DETOX is provided at \href{https://github.com/hwang595/DETOX}{https://github.com/hwang595/DETOX}.

\begin{figure*}[!h]
	\minipage{0.32\textwidth}
	\includegraphics[width=\linewidth]{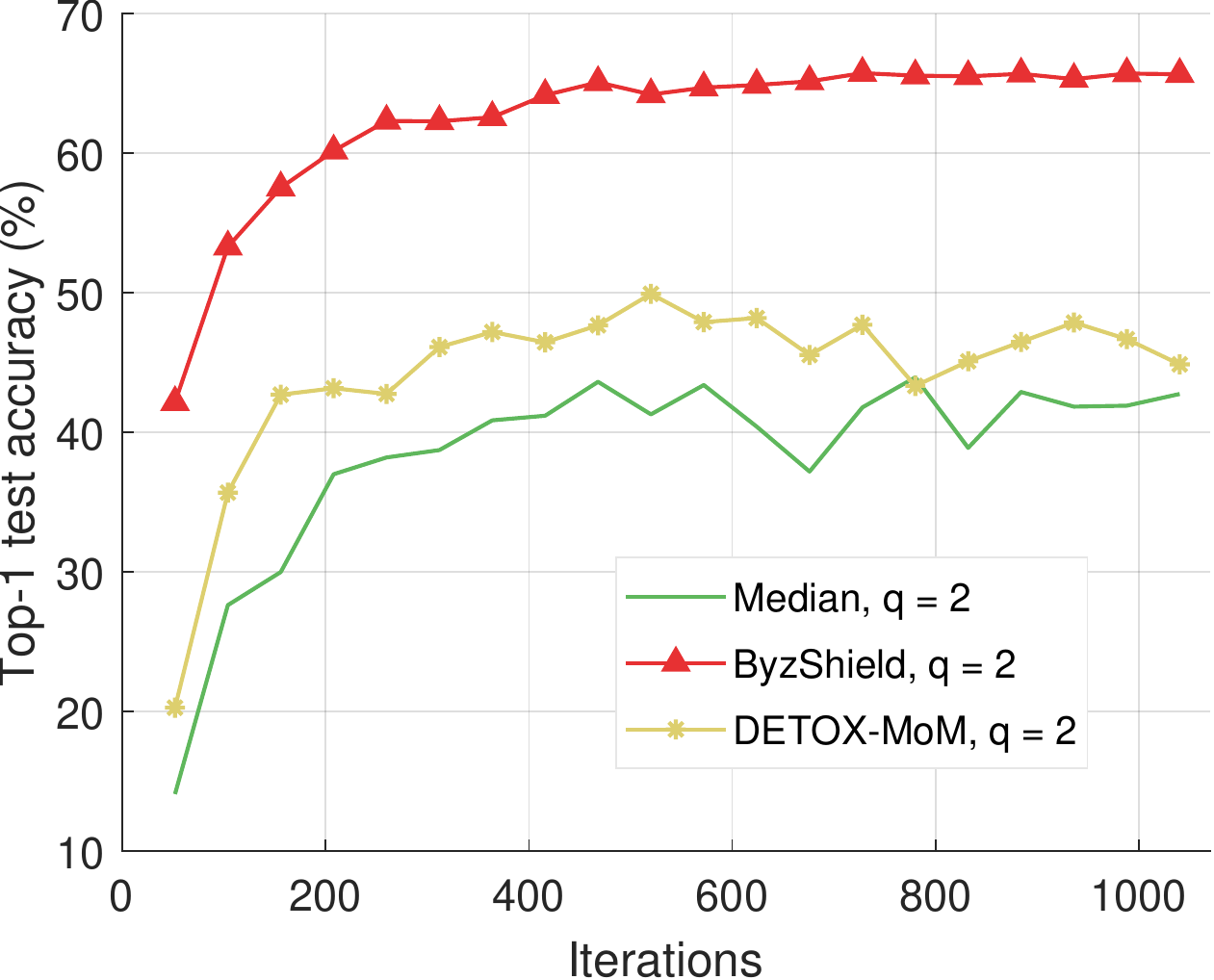}
	\caption{\emph{ALIE} attack and median-based defenses (CIFAR-10), $K=15$.}
	\label{fig:top1_fig_69}
	\endminipage\hfill
	\minipage{0.32\textwidth}
	\includegraphics[width=\linewidth]{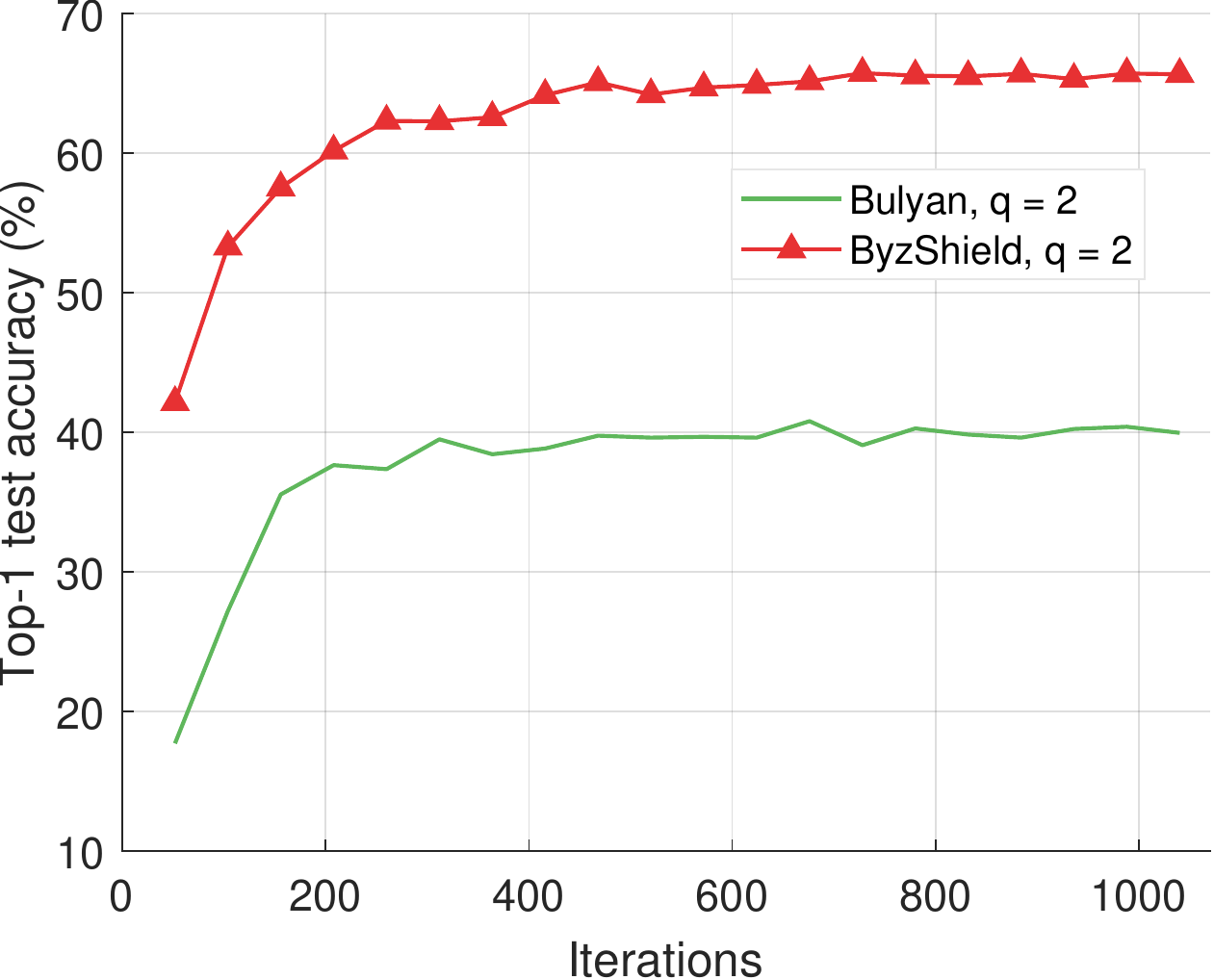}
	\caption{\emph{ALIE} attack and \emph{Bulyan}-based defenses (CIFAR-10), $K=15$}
	\label{fig:top1_fig_70}
	\endminipage\hfill
	\minipage{0.32\textwidth}
	\includegraphics[width=\linewidth]{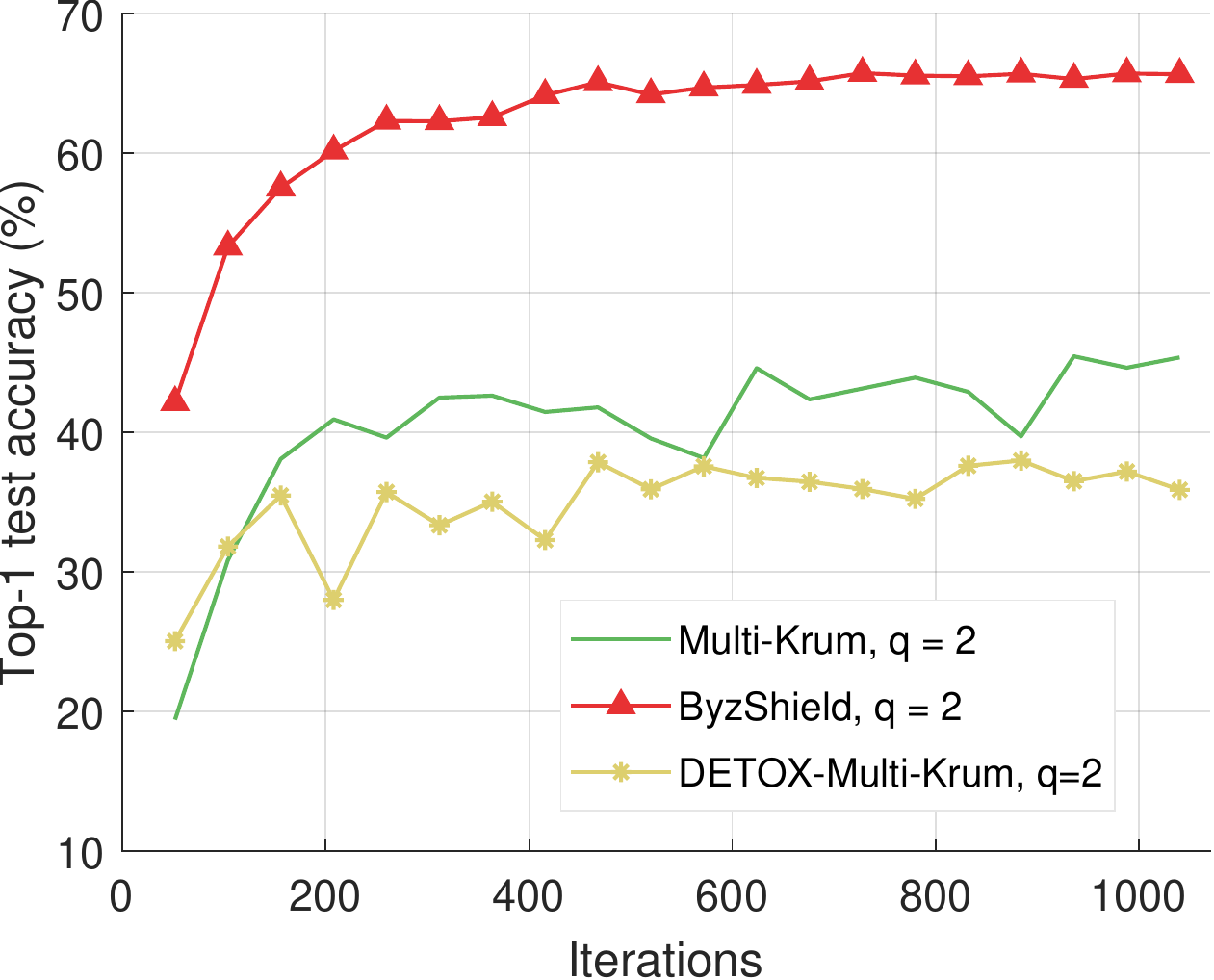}
	\caption{\emph{ALIE} attack and \emph{Multi-Krum}-based defenses (CIFAR-10), $K=15$.}
	\label{fig:top1_fig_71}
	\endminipage
	\vspace{-0.1in}
\end{figure*}

\begin{table}[t]
	\centering
	\caption{Distortion fraction evaluation for Ramanujan-based assignment (Case 2) for $(m,s)=(5,5)$, i.e., $(K,f,l,r)=(25,25,5,5)$ and comparison.}
	\label{table:Ramanujan_distortion_test_l5r5}
	%\resizebox{0.8\columnwidth}{!}{
	\setlength\tabcolsep{2pt}
	\begin{tabular}{ |P{0.6cm}||P{0.9cm}|P{1.8cm}|P{1.5cm}|P{1.1cm}|P{1.1cm}| }
		\hline
		$q$ & $c_{\mathrm{max}}^{(q)}$ & $\hat{\epsilon}^{ByzShield}$ & $\hat{\epsilon}^{Baseline}$ & $\hat{\epsilon}^{FRC}$ & $\gamma$ \\
		\hline
		$3$ & 1 & 0.04 & 0.12 & 0.2 & 2.43 \\
		$4$ & 1 & 0.04 & 0.16 & 0.2 & 3.9 \\
		$5$ & 2 & 0.08 & 0.2 & 0.2 & 5.56 \\
		$6$ & 4 & 0.16 & 0.24 & 0.4 & 7.35 \\
		$7$ & 5 & 0.2 & 0.28 & 0.4 & 9.25 \\
		$8$ & 7 & 0.28 & 0.32 & 0.4 & 11.23 \\
		$9$ & 9 & 0.36 & 0.36 & 0.6 & 13.28 \\
		$10$ & 12 & 0.48 & 0.4 & 0.6 & 15.38 \\
		$11$ & 14 & 0.56 & 0.44 & 0.6 & 17.54 \\
		$12$ & 17 & 0.68 & 0.48 & 0.8 & 19.73 \\
		\hline
	\end{tabular}
	%}
	%\vspace{-0.2in}
\end{table}

\begin{table}[t]
	\centering
	\caption{Distortion fraction evaluation for MOLS-based assignment for $(K,f,l,r)=(35,49,7,5)$ and comparison.}
	\label{table:MOLS_distortion_test_l7r5}
	%\resizebox{0.8\columnwidth}{!}{
	\setlength\tabcolsep{2pt}
	\begin{tabular}{ |P{0.6cm}||P{0.9cm}|P{1.8cm}|P{1.5cm}|P{1.1cm}|P{1.1cm}| }
		\hline
		$q$ & $c_{\mathrm{max}}^{(q)}$ & $\hat{\epsilon}^{ByzShield}$ & $\hat{\epsilon}^{Baseline}$ & $\hat{\epsilon}^{FRC}$ & $\gamma$ \\
		\hline
		$3$ & 1 & 0.02 & 0.12 & 0.14 & 2.68 \\
		$4$ & 1 & 0.02 & 0.16 & 0.14 & 4.39 \\
		$5$ & 2 & 0.04 & 0.2 & 0.14 & 6.36 \\
		$6$ & 4 & 0.08 & 0.24 & 0.29 & 8.54 \\
		$7$ & 5 & 0.1 & 0.28 & 0.29 & 10.89 \\
		$8$ & 8 & 0.16 & 0.32 & 0.29 & 13.37 \\
		$9$ & 10 & 0.2 & 0.36 & 0.43 & 15.97 \\
		$10$ & 11 & 0.22 & 0.4 & 0.43 & 18.67 \\
		$11$ & 14 & 0.29 & 0.44 & 0.43 & 21.44 \\
		$12$ & 16 & 0.33 & 0.48 & 0.57 & 24.29 \\
		$13$ & 20 & 0.41 & 0.52 & 0.57 & 27.2 \\
		\hline
	\end{tabular}
	%}
	%\vspace{-0.2in}
\end{table}

\newpage

\begin{table}[ht]
	\centering
	\caption{Distortion fraction evaluation for MOLS-based assignment for $(K,f,l,r)=(21,49,7,3)$ and comparison.}
	\label{table:MOLS_distortion_test_l7r3}
	%\resizebox{0.8\columnwidth}{!}{
	\setlength\tabcolsep{2pt}
	\begin{tabular}{ |P{0.6cm}||P{0.9cm}|P{1.8cm}|P{1.5cm}|P{1.1cm}|P{1.1cm}| }
		\hline
		$q$ & $c_{\mathrm{max}}^{(q)}$ & $\hat{\epsilon}^{ByzShield}$ & $\hat{\epsilon}^{Baseline}$ & $\hat{\epsilon}^{FRC}$ & $\gamma$ \\
		\hline
		$2$ & 1 & 0.02 & 0.1 & 0.14 & 2.23 \\
		$3$ & 3 & 0.06 & 0.14 & 0.14 & 4.67 \\
		$4$ & 5 & 0.1 & 0.19 & 0.29 & 7.72 \\
		$5$ & 8 & 0.16 & 0.24 & 0.29 & 11.29 \\
		$6$ & 12 & 0.24 & 0.29 & 0.43 & 15.27 \\
		$7$ & 16 & 0.33 & 0.33 & 0.43 & 19.6 \\
		$8$ & 21 & 0.43 & 0.38 & 0.57 & 24.22 \\
		$9$ & 25 & 0.51 & 0.43 & 0.57 & 29.08 \\
		$10$ & 29 & 0.59 & 0.52 & 0.71 & 34.15 \\
		\hline
	\end{tabular}
	%}
	%\vspace{-0.2in}
\end{table}

\begin{table}[!h]
	\centering
	\caption{Parameters used for training.}
	\label{table:tuning}
	%\resizebox{0.8\columnwidth}{!}{
	\begin{tabular}{ |P{0.9cm}||P{1.3cm}|P{3cm}| }
		\hline
		Figure & Schemes & Learning rate schedule \\
		\hline
		\ref{fig:top1_fig_62} & $1,2$ & $(0.00625,0.96,15)$\\
		\ref{fig:top1_fig_62} & $3$ & $(0.025,0.96,15)$\\
		\ref{fig:top1_fig_62} & $4,5,6$ & $(0.01,0.95,20)$\\
		\ref{fig:top1_fig_63} & $1,2$ & $(0.003125,0.96,15)$\\
		\ref{fig:top1_fig_64} & $1$ & $(0.00625,0.96,15)$\\
		\ref{fig:top1_fig_64} & $2,5,6$ & $(0.01,0.95,20)$\\
		\ref{fig:top1_fig_65} & $1,2$ & $(0.0001,0.99,20)$\\
		\ref{fig:top1_fig_65} & $3,4$ & $(0.025,0.96,15)$\\
		\ref{fig:top1_fig_65} & $5,6$ & $(0.001,0.5,50)$\\
		\ref{fig:top1_fig_66} & $1,2,4$ & $(0.05,0.96,15)$\\
		\ref{fig:top1_fig_66} & $3$ & $(0.1,0.95,50)$\\
		\ref{fig:top1_fig_66} & $5,6$ & $(0.025,0.96,15)$\\
		\ref{fig:top1_fig_67} & $1,2$ & $(0.025,0.96,15)$\\
		\ref{fig:top1_fig_67} & $4$ & $(0.05,0.96,15)$\\
		\ref{fig:top1_fig_68} & $1,2,3$ & $(0.05,0.96,15)$\\
		\ref{fig:top1_fig_68} & $7,8$ & $(0.025,0.96,15)$\\
		\ref{fig:top1_fig_69} & $1$ & $(0.003125,0.96,15)$\\
		\ref{fig:top1_fig_69} & $2$ & $(0.01,0.96,15)$\\
		\ref{fig:top1_fig_69} & $3$ & $(0.0125,0.96,15)$\\
		\ref{fig:top1_fig_70} & $1$ & $(0.0015625,0.96,15)$\\
		\ref{fig:top1_fig_71} & $1$ & $(0.003125,0.96,15)$\\
		\ref{fig:top1_fig_71} & $3$ & $(0.0125,0.96,15)$\\
		\hline
	\end{tabular}
	%}
	%\vspace{-0.2in}
\end{table}

\begin{figure}[!h]
	\centering
	\includegraphics[scale=0.5]{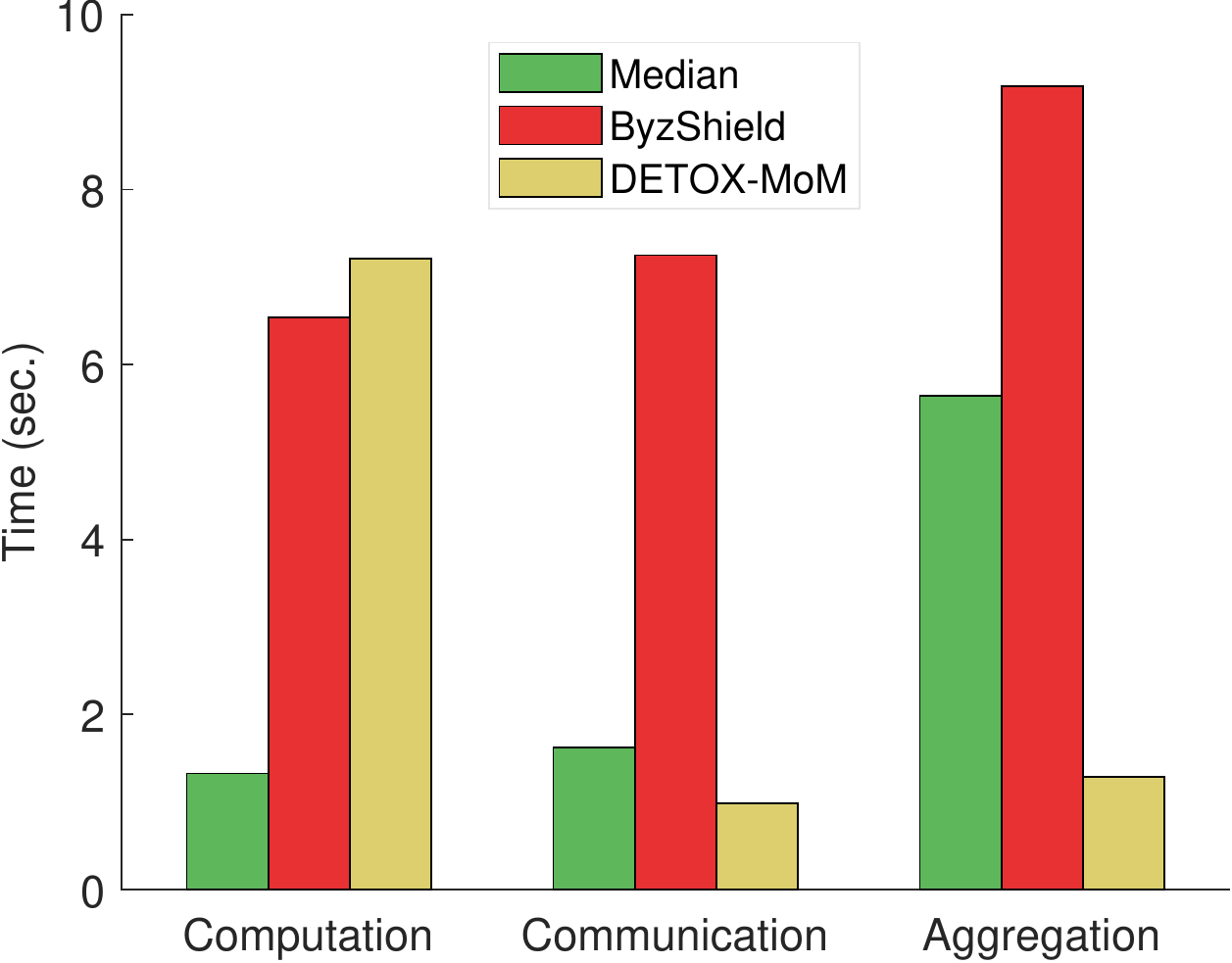}
	\caption{Per-iteration time estimate (\emph{ALIE} attack and median-based defenses (CIFAR-10)).}
	\label{fig:ALIE_median_per_iteration}
	\vspace{-0.1in}
\end{figure}

%%%%%%%%%%%%%%%%%%%%%%%%%%%%%%%%%%%%%%%%%%%%%%%%%%%%%%%%%%%%%%%%%%%%%%%%%%%%%%%
%%%%%%%%%%%%%%%%%%%%%%%%%%%%%%%%%%%%%%%%%%%%%%%%%%%%%%%%%%%%%%%%%%%%%%%%%%%%%%%

\end{document}